\documentclass{article}

\PassOptionsToPackage{super,comma,sort&compress}{natbib}

\usepackage[preprint]{neurips_2026}
\usepackage[utf8]{inputenc}
\usepackage[T1]{fontenc}
\usepackage{amsmath,amssymb}

\let\amsmathEqref\eqref

\usepackage{amsmath,amsfonts,bm}

\def\eqref#1{equation~\ref{#1}}

\def\1{\bm{1}}

\def\rvf{{\mathbf{f}}}
\def\rvg{{\mathbf{g}}}
\def\rvh{{\mathbf{h}}}

\def\rvr{{\mathbf{r}}}

\def\vzero{{\bm{0}}}

\def\va{{\bm{a}}}
\def\vb{{\bm{b}}}

\def\vs{{\bm{s}}}

\def\vu{{\bm{u}}}

\def\vx{{\bm{x}}}

\def\vz{{\bm{z}}}

\def\mI{{\bm{I}}}

\def\mL{{\bm{L}}}
\def\mM{{\bm{M}}}

\def\mX{{\bm{X}}}

\def\mZ{{\bm{Z}}}

\DeclareMathAlphabet{\mathsfit}{\encodingdefault}{\sfdefault}{m}{sl}
\SetMathAlphabet{\mathsfit}{bold}{\encodingdefault}{\sfdefault}{bx}{n}

\newcommand{\E}{\mathbb{E}}

\let\eqref\amsmathEqref
\usepackage{upgreek}

\newcommand{\rvPhi}{{\bm{\Phi}}}

\usepackage{caption}

\usepackage{needspace}
\usepackage{tikz}
\usetikzlibrary{arrows.meta,positioning,calc}
\usepackage{microtype,xurl,hyperref}

\makeatletter
\DeclareRobustCommand\citenum{\begingroup\NAT@swatrue\let\NAT@ctype\z@\NAT@parfalse
  \let\@cite\NAT@citenum\let\NAT@mbox\mbox\let\NAT@super@kern\relax\NAT@citexnum[][]}
  
\DeclareRobustCommand\citeref[1]{(\in@{,}{#1}\ifin@ refs\else ref.\fi~\citenum{#1})}
\makeatother

\usepackage{algorithm,algorithmic}

\newcommand*\samethanks[1][\value{footnote}]{\footnotemark[#1]}

\usepackage{xcolor}
\usepackage{booktabs,array,graphicx,adjustbox}
\usepackage{calc}
\usepackage{placeins,subcaption,wrapfig}
\usepackage{longtable}

\usepackage{caption}
\usepackage[nameinlink]{cleveref}
\crefname{figure}{Fig.}{Figs.}
\Crefname{figure}{Figure}{Figures}
\crefname{table}{Tab.}{Tabs.}
\Crefname{table}{Table}{Tables}
\crefname{subfigure}{Fig.}{Figs.}
\Crefname{subfigure}{Figure}{Figures}
\crefname{subtable}{Tab.}{Tabs.}
\Crefname{subtable}{Table}{Tables}
\crefname{suppfigure}{SFig.}{SFigs.}
\Crefname{suppfigure}{Supplementary Figure}{Supplementary Figures}
\crefname{supptable}{STab.}{STabs.}
\Crefname{supptable}{Supplementary Table}{Supplementary Tables}
\crefname{equation}{Eq.}{Eqs.}
\Crefname{equation}{Equation}{Equations}
\creflabelformat{equation}{#2#1#3}
\newcommand{\sectioncrefname}[1]{%
  \crefname{#1}{\textsection}{\textsection\textsection}%
  \Crefname{#1}{\textsection}{\textsection\textsection}%
  \crefformat{#1}{##2\textsection##1##3}%
  \Crefformat{#1}{##2\textsection##1##3}}
\AtBeginDocument{%
  \sectioncrefname{section}\sectioncrefname{subsection}\sectioncrefname{subsubsection}}
\crefname{appendix}{Appendix}{Appendices}
\Crefname{appendix}{Appendix}{Appendices}
\crefname{subappendix}{Appendix}{Appendices}
\Crefname{subappendix}{Appendix}{Appendices}
\crefname{subsubappendix}{Appendix}{Appendices}
\Crefname{subsubappendix}{Appendix}{Appendices}
\newcommand{\supplementaryfloats}{%
  \setcounter{figure}{0}\setcounter{table}{0}%
  \renewcommand{\figurename}{Supplementary Figure}%
  \renewcommand{\tablename}{Supplementary Table}%
  \crefalias{figure}{suppfigure}\crefalias{table}{supptable}%
  \crefalias{section}{appendix}\crefalias{subsection}{subappendix}%
  \crefalias{subsubsection}{subsubappendix}%
  \renewcommand{\theHfigure}{S\arabic{figure}}%
  \renewcommand{\theHtable}{S\arabic{table}}}
\graphicspath{{manuscript-assets/}{figures/output/04_paper_tables/}{figures/manual/}}

\definecolor{dclteal}{HTML}{359C94}
\definecolor{sigorange}{HTML}{DF8C29}
\definecolor{ink}{HTML}{293845}

\definecolor{stdgray}{gray}{0.45}
\newcommand{\stdpad}{}
\newcolumntype{P}[1]{>{\def\stdpad{#1}}r}
\newcommand{\meanmark}[1]{#1}
\newcommand{\meanstd}[2]{\ensuremath{{\meanmark{#1}}_{\textcolor{stdgray}{\text{\mdseries\sffamily\textpm\,\makebox[\maxof{\widthof{#2}}{\widthof{\stdpad}}][l]{#2}}}}}}
\newcommand{\won}[1]{{\bfseries\boldmath #1}}
\NewCommandCopy{\latexunderline}{\underline}
\makeatletter
\newcommand{\cellunderline}[1]{%
  \expandafter\ifx\@car#1\@nil\meanstd{\let\meanmark\latexunderline#1}\else\latexunderline{#1}\fi}
\makeatother
\AddToHook{env/tabular/begin}{\let\underline\cellunderline}
\AddToHook{env/tabular*/begin}{\let\underline\cellunderline}
\newcommand{\meanonly}[2][]{\ensuremath{{#2}_{\phantom{\text{\sffamily\textpm\,\stdpad}}}^{\text{#1}}}}

\hypersetup{hidelinks,pdftitle={world2vec: Stable inference of mechanistic world models},
  pdfauthor={Anonymous authors}}
\definecolor{aigreen}{HTML}{1A7F37}

\newcommand{\wv}{world2vec}
\newcommand{\WV}{World2vec}
\newcommand{\wvdcl}{\wv{}-dcl}
\newcommand{\wvsigreg}{\wv{}-sigreg}
\newcommand{\wvsindy}{\wv{}-SINDy}
\newcommand{\Wvdcl}{\WV{}-dcl}
\newcommand{\Wvsigreg}{\WV{}-sigreg}
\newcommand{\Wvsindy}{\WV{}-SINDy}
\makeatletter
\renewcommand{\paragraph}{\@startsection{paragraph}{4}{\z@}{\z@}%
  {-\fontdimen2\font \@plus -\fontdimen3\font \@minus -\fontdimen4\font}%
  {\normalfont\normalsize\bfseries}}
\makeatother
\title{world2vec: Mechanistic world models \\ through latent dynamics learning}
\author{
    Tobias Schmidt\thanks{%
        Equal contribution
        $\qquad ^\dagger$ Correspondence to \texttt{steffen.schneider@helmholtz-munich.de}
    }, \ 
    Rodrigo Gonzalez Laiz\samethanks{}, \ 
    Krisztián Szuppinger %
    \& Steffen Schneider$^\dagger$\\[.25em]
    Helmholtz Munich \& Munich Center for Machine Learning (MCML)
}

\begin{document}
\maketitle

\begin{abstract}
    World models can achieve strong control performance without faithfully capturing the dynamics of their environment, limiting their potential for mechanistic interpretability and scientific discovery. 
    Here, we introduce \wv{} to learn mechanistic world models without reward information directly from observations and actions. %
    \WV{} learns latent dynamics that faithfully reproduce the true system over long prediction horizons, with at least one order of magnitude fewer parameters than DreamerV3, TD-MPC2 and LeWM. This enables state of the art performance in controlling continuous environments from pixels and accurate system identification across a benchmark of 71 attractors, demonstrating how world models are amenable to scientific discovery.
    \\
    \begin{center}
        \vspace{-1em}
        \url{https://worldmodels.science}
    \end{center}
\end{abstract}

\section{Introduction}\label{introduction}

    Estimating models of dynamically changing systems and environments is an important problem at the frontier of representation learning, with applications in physical AI and scientific discovery\citep{may1976simple,ha2018world,ha2018recurrent,course2023state,mathis2026joint,lecun2022jepa,schmidhuber1990making}.
    Here we focus on the problem where a state $\vs$ evolves under actions $\va$ according to dynamics $\rvf$, and we observe this state through a non-linear mixing function $\rvg$:
    \begin{equation}\label{eq:system}
        \dot{\vs} = \rvf(\vs, \va), \qquad \vx = \rvg(\vs).
    \end{equation} 
    Then, latent world modeling\citep{lecun2022jepa} is the problem of jointly estimating, from observations and actions alone, an encoder $\rvh$ that maps observations to a representation $\vz=\rvh(\vx)$ and a latent dynamics model $\hat\rvf$ that describes how this representation evolves in time.
    Ideally, the representation is a change of coordinates of the state, $\vz=\rvr(\vs)$ with $\rvr=\rvh\circ\rvg$ invertible, and the latent dynamics reproduce the true evolution in these coordinates\citep{laiz2026understanding},
    \begin{equation}\label{eq:wm-goal}
        \dot{\vz}=\hat\rvf(\vz,\va),\quad \vz(0)=\rvr(\vs(0)) 
        \quad\Longrightarrow\quad
        \vz(t)=\rvr(\vs(t))\ \ \text{for all } t\geq0.
    \end{equation}
    If this condition holds, the world model $(\rvh, \hat \rvf)$ can support model-predictive control or model-based reinforcement learning to steer the environment into desired target states.
    Over the past years, this type of world modeling has become feasible and gained theoretical grounding. Common objectives are reconstruction-based forecasting\citep{dreamer}, reconstruction-free planning in latent space, either trained with rewards\citep{burchi2025learning,tdmpc2} or reward-free using the joint-embedding predictive architecture\citep{lecun2022jepa,lewm,zhou2024dino} (JEPA).
    
    \begin{figure}[t]
        \centering
        \includegraphics[width=\textwidth]{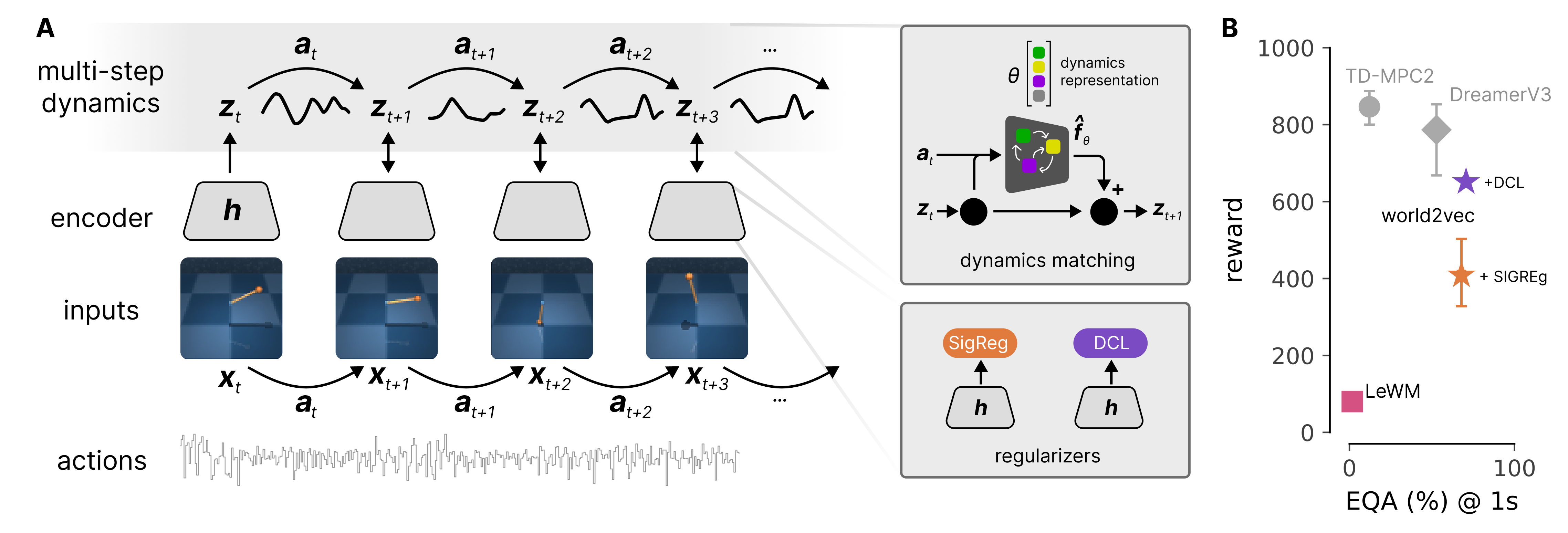}
        \caption{%
            \textbf{A.}~\WV{} jointly learns representations and dynamics in latent space; the dynamics model is an action-conditioned neural ODE which can be parameterized to provide an interpretable vector representation of the dynamics. The objective function incorporates assumptions about the underlying data-generating process.
            \textbf{B.}~Reward vs.\ 1\,s Equivariance Accuracy (EQA) of \wv{} vs.\ baselines, averaged across environments. %
            }
            \vspace{-0.5em}
        \label{fig:framework}
    \end{figure}
    
    Importantly, world modeling extends beyond solving a particular control task. We also seek to ensure that the world model accurately reflects the world in a statistical sense: Does the learned dynamics model reproduce the relevant dynamics when the statistical assumptions used to build it are satisfied?
    If satisfied, an appropriate parameterization can further enable mechanistic interpretability by allowing us to read out the environmental dynamics.
    This has value both for interpretability and scientific applications. World models with these properties have recently been characterized as \emph{mechanistic} world models\citep{mech_world_models}.
    Building on ideas from non-linear independent component analysis\citep{pmlr-v89-hyvarinen19a}, it has been shown that self-supervised learning is able to provably estimate latent representations and their dynamics from data \citep{dcl} across a range of objective functions. \citep{mikulasch2026understanding,mikulasch2026predictive}
    Transferring these theoretical guarantees into algorithms that scale to challenging control problems is the goal of this present work. 
    
    In this paper, we propose \emph{\wv{}} (\cref{fig:framework}A) to advance the shared goal of building a world model that
        (\emph{i}) achieves state-of-the-art performance on control tasks after training on offline data alone,
        (\emph{ii}) faithfully models the environment, measured via long-range rollouts and predictions, and
        (\emph{iii}) is mechanistically interpretable and thus suitable for scientific applications.
    \WV{} is agnostic to its exact training objective, provided it belongs to a class of objectives known to yield identifiable representations under idealized conditions. We consider LeJEPA \citep{lejepa,lewm} and dynamics contrastive learning\citep{dcl} (DCL) as two possible approaches to achieving this goal. These objectives encode statistical assumptions with different implications on numerical estimation, and an increasing body of theoretical work addresses the identifiability of these objectives \citep{zimmermann_contrastive_2021,dcl,lejepa_identifiability, mikulasch2026understanding}.
    
    Across three considered environments, \wv{} jointly achieves high reward and performance in forecasting (\cref{fig:framework}B), enabling successful model predictive control in 2 of 3 environments, none of which were solvable by the previous best JEPA-based approach\citep{lewm}.
    In the remainder of the paper, we introduce the idea of \emph{dynamics matching} and the \wv{} framework (\cref{sec:towards-mechanistic-world-models}), which we benchmark on control across several MuJoCo environments (\cref{sec:setup}) and on symbolic regression (\cref{sec:sindy}).

\section{The world2vec framework for mechanistic world modeling}\label{sec:towards-mechanistic-world-models}

    Mechanistic world modeling\citep{mech_world_models} seeks to recover a representation of the physical state, a faithful model of its evolution, and ideally an interpretable description of the governing equations.
    For the system in \cref{eq:system}, we observe $\vx$ and $\va$, but have no direct access to $\vs$ or $\rvf$.
    We aim to learn an encoder $\rvh$ that maps observations to latent states $\vz=\rvh(\vx)$, and their dynamics $\dot{\vz}=\hat\rvf(\vz,\va)$.

    \paragraph{Faithful latent dynamics.}
    When does this learned model faithfully describe the physical system?
    Let $\rvr=\rvh\circ\rvg$ denote the mapping from physical to latent states.
    If $\rvr$ is invertible, the latent representation is simply a change of coordinates of the physical state.
    The model is then faithful\citep{laiz2026understanding} if encoding the evolved physical state yields the same trajectory as evolving the encoded state in latent space (\cref{eq:wm-goal}).
    Integrating the latent dynamics over a horizon $T$, we require:
    \begin{equation}\label{eq:faithful}
        \hat\vz_t(T) := \vz(t)+\int_0^T\hat\rvf\big(\hat\vz_t(\tau),\va(t+\tau)\big)\,d\tau
        \overset{!}{=}
        \rvh\big(\vx(t+T)\big)
        \qquad\text{for all } t,\ T\geq0.
    \end{equation}
    Importantly, both sides of this condition can be computed from observations and actions alone, without access to the states $\vs$ or the ground truth dynamics $\rvf$.

\paragraph{The \wv{} objective.}
    To learn the encoder--dynamics pair $(\rvh,\hat\rvf)$ in practice, we collect a dataset $\mathcal D$ with observations sampled at intervals $\Delta t$, i.e., at times $t_n=n\Delta t$, and recorded actions.
    Under partial observability, a single observation does not necessarily determine the state; for example, velocities cannot be inferred from a single frame.
    The encoder therefore takes a short history $\mX_n=(\vx_{n-H+1},\ldots,\vx_n)$ of length $H$, chosen such that the state can be recovered from it, i.e., such that the map from $\vs(t_n)$ to $\mX_n$ is injective\citep{dcl}, and maps it to the latent state $\vz_n=\rvh(\mX_n)$.
    For training, we sample starting points $n\sim p_{\mathcal D}$ and select observations $\vx_{n-H+1:n+K}$ and actions $\va_{n:n+K-1}$ so we can roll out the model for $K$ steps.

    The dynamics model $\hat\rvf(\vz,\va)$ predicts the latent rate of change, which a numerical integration step $\rvPhi$ uses to advance the state by $\Delta t$.
    We hold each action constant within its sampling interval and obtain the latent rollout
    \begin{equation}\label{eq:recursion}
        \hat\vz_{n,0}=\rvh(\mX_n),\qquad
        \hat\vz_{n,j+1}=\rvPhi(\hat\vz_{n,j},\va_{n+j}),
    \end{equation}
    where $\hat\vz_{n,j}$ is the inferred latent state predicted $j$ steps ahead from starting point $n$, to be compared to the encoded observation $\vz_{n+j}=\rvh(\mX_{n+j})$.
    Our main experiments use explicit Euler integration:
    \begin{equation}\label{eq:euler}
        \rvPhi(\vz,\va)=\vz+\Delta t\,\hat\rvf(\vz,\va).
    \end{equation}
    Evaluating \cref{eq:faithful} at $T=j\Delta t$ for $j=1,\ldots,K$ then gives the training objective
    \begin{equation}\label{eq:rollout}
        \min_{\rvh,\hat\rvf}\;
        \E_{n\sim p_{\mathcal D}}\!\left[\sum_{j=1}^K w_j\left\|\rvh(\mX_{n+j})-\hat\vz_{n,j}\right\|_2^2\right]
        +\lambda\mathcal R(\rvh,\hat\rvf)
        \quad\text{\emph{(world2vec objective)}},
    \end{equation}
    where $w_j$ are optional weightings for each step and $\mathcal{R}(\rvh, \hat \rvf)$ is a regularizer, weighted by $\lambda\geq0$, to prevent representational collapse (a constant encoder trivially satisfies \cref{eq:faithful}).
    In dynamics contrastive learning (DCL), prediction scores based on squared distances are normalized against sampled negative embeddings \citep{dcl}.
    In the LeJEPA instantiation, SIGReg regularizes embedding projections toward an isotropic Gaussian \citep{lejepa}.
    We refer to the resulting models as \emph{\wvdcl{}} and \emph{\wvsigreg{}} and give details on the full objectives in \cref{app:objectives}.

    \paragraph{From single-step to multi-step dynamics matching.}
    \Cref{eq:faithful} is a condition on ``latent velocities''. Integrated along the latent rollout, the learned velocity field $\hat\rvf$ must reproduce the change of the encoded state $\vz(t+T)-\vz(t)$.
    We term this principle \emph{latent dynamics matching}.
    For a single step ($K=1$), the \wv{} objective matches velocities directly, up to a finite-difference approximation:
    \begin{align}\label{eq:ode}
      \left\|\dot{\vz}(t_n)-\hat \rvf(\vz_n,\va_n)\right\|_2^2
      \approx\left\|\frac{\vz_{n+1}-\vz_n}{\Delta t}
      -\hat\rvf(\vz_n,\va_n)\right\|_2^2
        \propto \left\| \left[ \vz_n + \Delta t\,\hat\rvf(\vz_n,\va_n) \right] - \vz_{n+1} \right\|_2^2,
    \end{align}
    where the right-hand side is the single-step term of \cref{eq:rollout}, computed with the residual predictor of \cref{eq:euler}, similar to the approach in Neural ODEs\citep{chen2018neural}.
    Over multiple steps, the terms of \cref{eq:rollout} match the encoded displacement with the velocities accumulated along the model's predicted trajectory:
    \begin{align}\label{eq:multistep}
      \left\|\frac{\vz_{n+j}-\vz_n}{\Delta t}
      -\sum_{\ell=0}^{j-1}\hat\rvf(\hat\vz_{n,\ell},\va_{n+\ell})\right\|_2^2
        \propto \left\| \hat\vz_{n,j} - \vz_{n+j} \right\|_2^2.
    \end{align}
    Training on this residual exposes how small one-step errors compound over long rollouts.
    We hypothesize that modeling state \emph{change} through a learned velocity field provides a useful inductive bias for world modeling.

\paragraph{Empirical identifiability of latents and their dynamics.}
    The world2vec algorithm is a numerical implementation of learning principles that have been extensively studied theoretically and are known to produce representations that are consistent up to at least affine transformations \citep{zimmermann_contrastive_2021,mikulasch2026understanding,lejepa_identifiability}.
    The challenge lies in the numerical estimation to obtain this guarantee in practice. As an empirical measure of identifiability, we check whether independently trained encoders recover equivalent representations up to an affine change of coordinates, termed consistency\citep{schneider2023cebra}.
    To estimate consistency, we report the $R^2$ after optimal affine alignment for the two encoders $\rvh_1$ and $\rvh_2$,
    \begin{equation}\label{eq:embedding_alignment}
        \max_{\mL, \vb}
           R^2\left[\{\rvh_2(\mX_n)\}_n,\{\mL\rvh_1(\mX_n)+\vb\}_n\right],
    \end{equation}

    across independent training runs. Within each model, we also assess whether the learned dynamics reproduce the evolution of its representation using Latent Rollout Consistency (LRC)\citep{laiz2026understanding},
    \begin{equation}\label{eq:rollout_equivariance}
     \mathrm{LRC}_j
     = R^2\!\left[
       \{\rvh(\mX_{n+j})\}_{n},
       \{\hat{\vz}_{n,j}\}_{n}
     \right],
    \end{equation}
    where $j$ is the prediction horizon. This score compares predictions and encoded futures from the same encoder--dynamics pair, but can be prone to collapse (i.e., a trivial embedding could get a high LRC).
    Hence, we also adopt the accuracy-based equivariance measure of \citet{schmidt2025ebc} which for our time series case we denote as Equivariance Accuracy (EQA).
    EQA is the top-1 retrieval accuracy of the encoded future observation among negative candidates.
    We select model hyperparameters based on validation EQA and use 512 negatives in all experiments.

\paragraph{Dynamics parameterization.}
    The function class for $\hat{\rvf}$ can be flexibly chosen within the world2vec framework. For our main experiments focused on control, we utilize an MLP that maps latent state and action to the rate of change.
    As a more interpretable parameterization, we build on SINDy\citep{sindy} and the more recent DYSCO\citep{dysco} and fit the structured dynamics model
    \begin{equation}\label{eq:library}
     \hat{\rvf}(\vz,\va)=\boldsymbol{\Xi}\,\boldsymbol{\theta}(\vz,\va),
    \end{equation}
    where $\boldsymbol{\theta}(\vz,\va)\in\mathbb{R}^{p}$ is a fixed vector of library features and $\boldsymbol{\Xi}\in\mathbb{R}^{d\times p}$ contains learned coefficients for a $d$-dimensional latent space. Our SINDy instantiation uses polynomial features for $\boldsymbol{\theta}$ and an $L_1$ penalty on $\boldsymbol{\Xi}$ \citep{sindy,sindy_ae,sindy_bae}.

\paragraph{Model-predictive control.}
    Given a sequence of target frames $\mX^\star$, we define $\vz^\star=\rvh(\mX^\star)$ as the goal state.
    Starting from the current history $\mX_n$, we sample candidate action sequences of planning horizon $P$ with the cross-entropy method\citep{rubinstein1999cross} (CEM) using the rollout in \cref{eq:recursion} and minimize
    \begin{equation}\label{eq:planning}
     \min_{\va_{n:n+P-1}} \sum_{j=1}^P\beta_j\|\hat{\vz}_{n,j}-\vz^\star\|_2^2,
     \qquad \text{ s.t. }\beta_j\geq0,\quad\sum_{j=1}^P\beta_j=1,
    \end{equation}
    subject to the environment's action bounds.
    The controller executes a prefix of the selected plan and replans from newly observed data, using neither a learned reward model nor a state decoder.

\section{Faithful dynamics modeling for continuous control}\label{sec:setup}

\subsection{Experimental Setup}

\begin{wrapfigure}{r}{0.55\textwidth}
    \vspace{-1em}
    \centering
    \includegraphics[width=\linewidth]{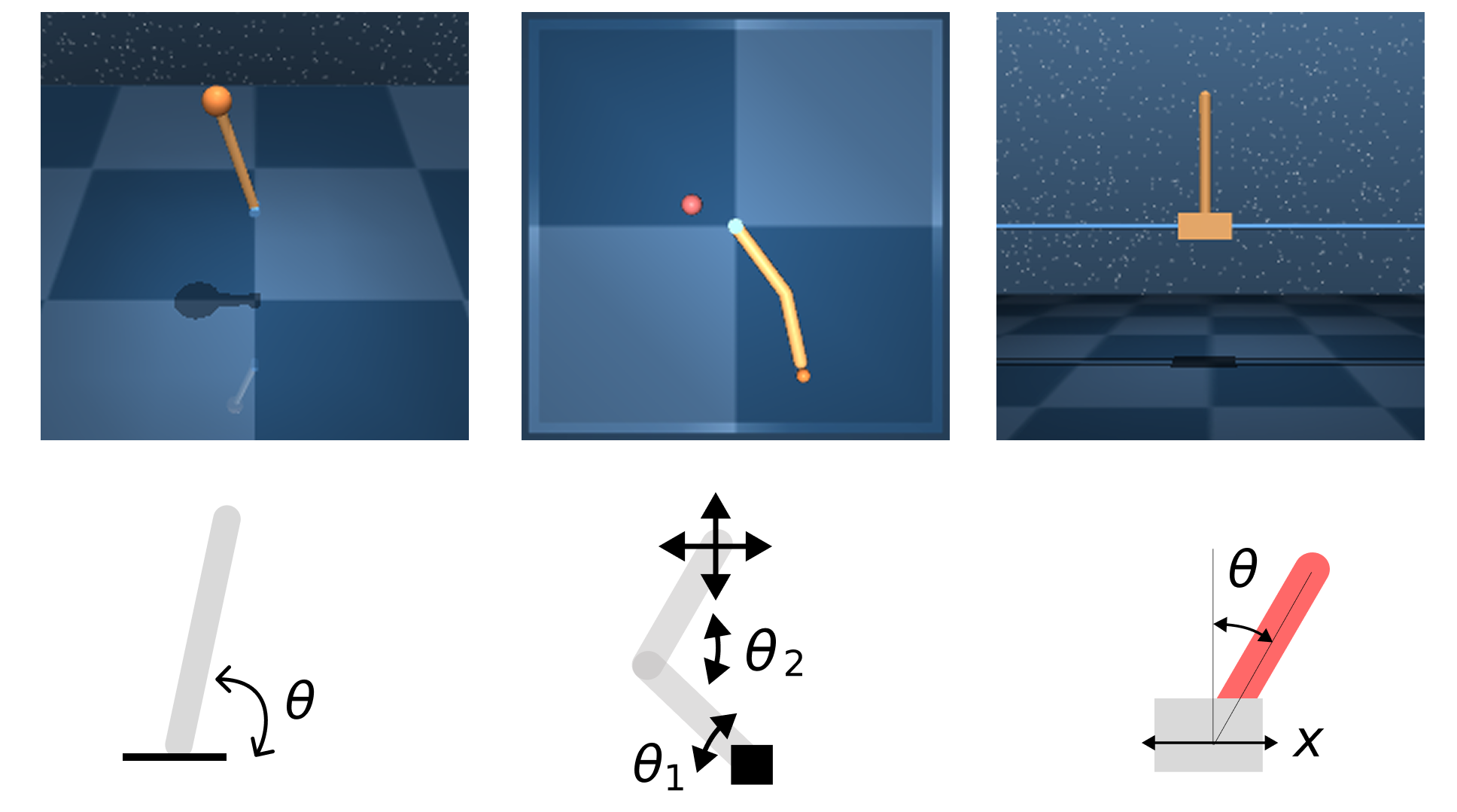}
    \caption{Pendulum, Reacher and CartPole environments used in our study with their physical variables.}
    \label{fig:data}
    \vspace{-1em}
\end{wrapfigure}
\paragraph{Environments and dataset collection.}
    We evaluate \wv{} on the Pendulum, Reacher, and Cartpole environments from the MuJoCo-based DeepMind Control Suite\citep{control_suite} (\cref{fig:data}).
    We collect 200k RGB frames with 64$\times$64\,px resolution using random policies in each environment: uniform random actions for Reacher, and temporally correlated Ornstein--Uhlenbeck actions for Pendulum and Cartpole (details in \cref{app:dataset_collection}).
    For identifiability analysis, we construct a variant of this dataset with non-visual, non-linear observations.
    Following \citet{dcl}, the positional and angular coordinates (but not  velocities) are passed through a randomly initialized, injective MLP of varying depth and projected to 50 dimensions.

\paragraph{Model families and training.}
    Encoders take a short sequence of $H=3$ observations as input.
    A small vision transformer encodes each frame (16$\times$16 patches, 2 layers of width 64 with 4 heads, and a keypoint readout) and is followed by an MLP head (2 hidden layers of 128 units) that maps the sequence of frame representations to the latent state.
    On nonlinear-mixing observations, we concatenate the observation sequence and pass it through an MLP encoder (3 hidden layers of 128 units).
    Dynamics models are MLPs with 3 hidden layers of 128 units. To model continuous-time dynamics, the MLP parametrizes the vector field $\hat\rvf$ and $\rvPhi$ is one Euler step (\cref{eq:euler}).
    As a baseline close to existing JEPA-setups, we parameterize discrete-time dynamics by replacing the Euler step (\cref{eq:euler}) with $\rvPhi(\vz,\va)=\mathrm{MLP}(\vz,\va)$, where the MLP directly predicts the next latent state. 
    With a latent dimensionality of $d=32$, the main \wv{} model has 263k parameters (221k for the encoder, 42k for the dynamics).
    All models are trained for 20k iterations with Adam using learning rate $10^{-3}$ and cosine decay. 

\paragraph{Model predictive control.}
    We control the environments with model predictive control (MPC) in latent space without access to reward.
    Instead, we assume knowledge of an observation sequence that shows the desired target. We encode this sequence to obtain a goal latent that describes which part of the latent space the environment should be driven towards. For Pendulum and Cartpole, this is a sequence of frames that show the pole or pendulum upright at rest, with the cart centered. For Reacher, the target changes in every episode, and the goal is underspecified since two arm configurations place the fingertip on the target. We therefore compute, per episode, the arm configuration that places the fingertip on the target via inverse kinematics, render it, and use its encoding as the goal. Both possible elbow configurations are planned and the average between them is reported.
    We then sample candidate action sequences using the cross-entropy method\cite{rubinstein1999cross} (CEM), predict future latent states with the dynamics model and rank them via their distance to the goal state (\cref{eq:planning}). We execute the first actions of the best sequence, observe the environment, and plan again. Because the continuous-time dynamics can be integrated with a different step size, it is possible to use different sampling rates for training and testing.

\paragraph{Baselines and Metrics.}
    TD-MPC2 and DreamerV3 are trained online following their respective training protocols. LeWM and \wv{} are trained offline on a random-policy dataset\cref{app:dataset_collection}. For Reacher, we report LeWM trained on the original LeWM dataset\citep{lewm} in the main results and compare it with LeWM trained on our dataset in \cref{tab:lewm-protocol}. All models are evaluated on the same held-out test episodes of our datasets.
    As a controlled baseline within our own framework, we implement a LeWM variant using discrete dynamics, one-step prediction, and the SIGReg regularizer. This baseline has similar performance to LeWM in terms of LRC and EQA and is substantially faster due to a more compact model.
    We measure the quality of the representation through consistency and the quality of the rollout using LRC and EQA as detailed in \cref{sec:towards-mechanistic-world-models}.
    Hyperparameters are selected using the highest mean validation EQA on the validation split.
    We measure task performance via cumulative reward. All metrics are outlined in detail in \cref{app:metrics}.
    
\paragraph{Implementation.}
    We implemented \wv{} in PyTorch. For the considered control tasks, small models with 64$\times$64\,px frame inputs are sufficient, while the framework allows encoders with larger inputs. On a single NVIDIA A100 GPU, training takes only 2--6 minutes for a one-step model, about 10 minutes for a 20-step model, and about 30 minutes for a 40-step model.
    Closed-loop control takes about 20--50\,ms per control step, including simulation and rendering.
    Beyond empirical performance, this efficiency enables downstream applications in robotics and closed-loop experimental settings.

\subsection{Results}\label{sec:results}

\begin{table}[t]
    \caption{%
    Consistency, LRC and EQA (\%) on the test split, averaged over environments, and reward
    per environment (maximum 1000; best planner per environment). Mean $\pm$ bootstrap standard error over model seeds; best in bold, second-best underlined.
    On Reacher, the baselines and \wv{} run at 8.3\,Hz, TD-MPC2 and DreamerV3 at 25\,Hz, and LeWM, trained on the
    original LeWM data, at 5\,Hz (see \cref{app:baselines} \cref{tab:configurations-won,tab:controllers} for details on hyperparameters).}
    \label{tab:main}
    \small\centering
    \begin{adjustbox}{max width=\linewidth}
\begin{tabular*}{1.03\linewidth}{@{\extracolsep{\fill}}lrcccccc@{}}\toprule
& & & & & \multicolumn{3}{c}{Reward} \\
\cmidrule(lr){6-8}
Method & \# params & Cons. & LRC$_{1\,\mathrm{s}}$ & EQA$_{1\,\mathrm{s}}$ & Pendulum & Reacher & CartPole \\
\midrule
DreamerV3 \citeref{dreamer} & 17M & 43.4 & \meanstd{52.3}{1.4} & \meanstd{52.7}{0.5} & \meanstd{668}{140} & \underline{\meanstd{792}{25}} & \won{\meanstd{899}{6}} \\
TD-MPC2 \citeref{tdmpc2} & 5M & 87.5 & \meanstd{15.5}{2.4} & \meanstd{12.1}{0.5} & \won{\meanstd{911}{2}} & \meanstd{744}{41} & \underline{\meanstd{894}{9}} \\
LeWM \citeref{lewm} & 18M & \won{99.5} & \meanstd{0.0}{0.0} & \meanstd{1.7}{0.4} & \meanstd{5}{0} & \meanstd{69}{4} & \meanstd{171}{5} \\
\midrule
Baseline (SIGReg $\approx$ LeWM) & 300k & 77.4 & \meanstd{37.4}{1.7} & \meanstd{20.3}{1.2} & \meanstd{12}{10} & \meanstd{87}{6} & \meanstd{201}{4} \\
Baseline (DCL) & 300k & 88.3 & \meanstd{51.2}{0.8} & \meanstd{11.1}{0.2} & \meanstd{3}{3} & \meanstd{104}{8} & \meanstd{154}{8} \\
\wvsigreg{} & 300k & \underline{96.9} & \underline{\meanstd{96.1}{0.0}} & \underline{\meanstd{67.9}{1.0}} & \meanstd{434}{142} & \meanstd{570}{20} & \meanstd{221}{1} \\
\wvdcl{} & 300k & 95.5 & \won{\meanstd{97.7}{0.1}} & \won{\meanstd{70.7}{0.7}} & \underline{\meanstd{826}{0}} & \won{\meanstd{895}{16}} & \meanstd{232}{5} \\
\bottomrule
\end{tabular*}
\end{adjustbox}

\end{table}

\begin{figure}[t]
    \centering
    \vspace{-0.75em}
    \includegraphics[width=\linewidth]{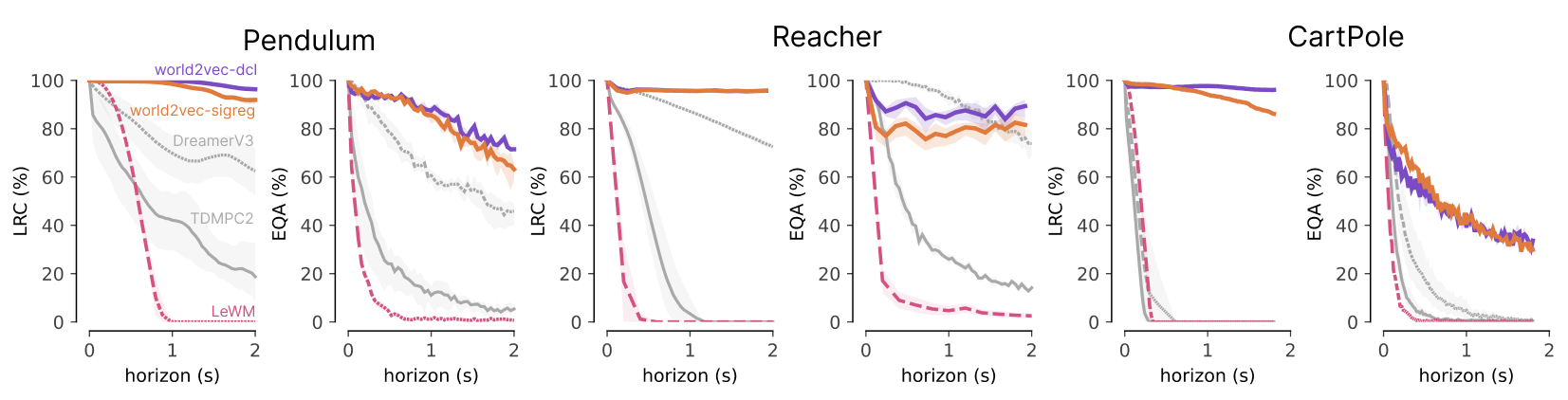}\\[-0.35em]
    \caption{%
        LRC and EQA vs.\ rollout horizon across environments.
        Models and selection as in \cref{tab:main}.
        }
    \label{fig:main-rollouts}
    \vspace{-0.5em}
\end{figure} 

\paragraph{\WV{} faithfully models dynamics and enables long-range predictions and control.}

\WV{} improves latent prediction substantially, enabling successful model predictive control in 2 of the 3 environments, none of which were solvable by the previous best JEPA-based approach, for which rewards remain low on all three tasks (\cref{fig:framework}B; \cref{tab:main}; \cref{fig:main-rollouts}). Combining multistep prediction ($K \in [10,20,40]$) with an Euler-integrated latent vector field raises 1-second LRC from 37--51\% in our one-step baselines to 96--98\%, and mean rewards from chance to 434--826 on Pendulum and 570--895 on Reacher (\cref{tab:main}). At a 30-step horizon the same LRC comparison is 0\% versus 73--86\% (\cref{fig:main-rollouts}).
The gains in rollout consistency are most noticeable for Pendulum and CartPole (\cref{fig:main-rollouts}), which have substantial intrinsic dynamics.
Reward-based models attain high rewards on all three tasks but have poor latent prediction, except on Reacher (\cref{fig:framework}B, gray).
As predicted by theory on linear identifiability\citep{roeder2021linear}, our results also demonstrate that latent world models learn representations that remain consistent across training runs. \WV{} and LeWM obtain an average consistency score of > $95\%$ in contrast to $87.5\%$ in TD-MPC2 and $43.4\%$ in DreamerV3.

    \begin{wrapfigure}{r}{0.52\textwidth}
        \vspace{-1.25em}
        \centering
        \captionof{table}{Reacher under the LeWM goal-reaching protocol: success rate (\%) for goals 25 or 200 steps ahead (budget kept at 50 steps), and planning time per episode (s/ep., for goals 25 steps ahead). Mean $\pm$ bootstrap standard error over training seeds; best in bold, second-best underlined. Full results in \cref{tab:lewm-protocol}.}
        \label{tab:reacher-timing}
        \footnotesize
        \setlength{\aboverulesep}{0.25ex}\setlength{\belowrulesep}{0.4ex}
        \setlength{\tabcolsep}{3pt}
\begin{tabular*}{\linewidth}{@{\extracolsep{\fill}}lccc@{}}
\toprule
 & \multicolumn{2}{c}{Success} & \\
\cmidrule(lr){2-3}
Method & 25 steps & 200 steps & s/ep. \\
\midrule
LeWM & \meanstd{81.6}{2.4} & \meanstd{52.9}{3.8} & \meanstd{8.64}{0.12} \\
\wvsigreg{} 25\,Hz & \underline{\meanstd{85.8}{1.0}} & \meanstd{72.4}{0.2} & \meanstd{2.61}{0.06} \\
\wvdcl{} 25\,Hz & \won{\meanstd{86.2}{1.1}} & \won{\meanstd{80.2}{1.6}} & \meanstd{2.55}{0.03} \\
\wvsigreg{} 8.3\,Hz & \meanstd{78.9}{1.4} & \meanstd{76.0}{2.2} & \underline{\meanstd{1.08}{0.07}} \\
\wvdcl{} 8.3\,Hz & \meanstd{75.8}{2.7} & \underline{\meanstd{77.6}{2.8}} & \won{\meanstd{1.00}{0.01}} \\
\bottomrule
\end{tabular*}

        \vspace{-1em}
    \end{wrapfigure}
\begin{figure}[t]\centering
    \includegraphics[width=\linewidth]{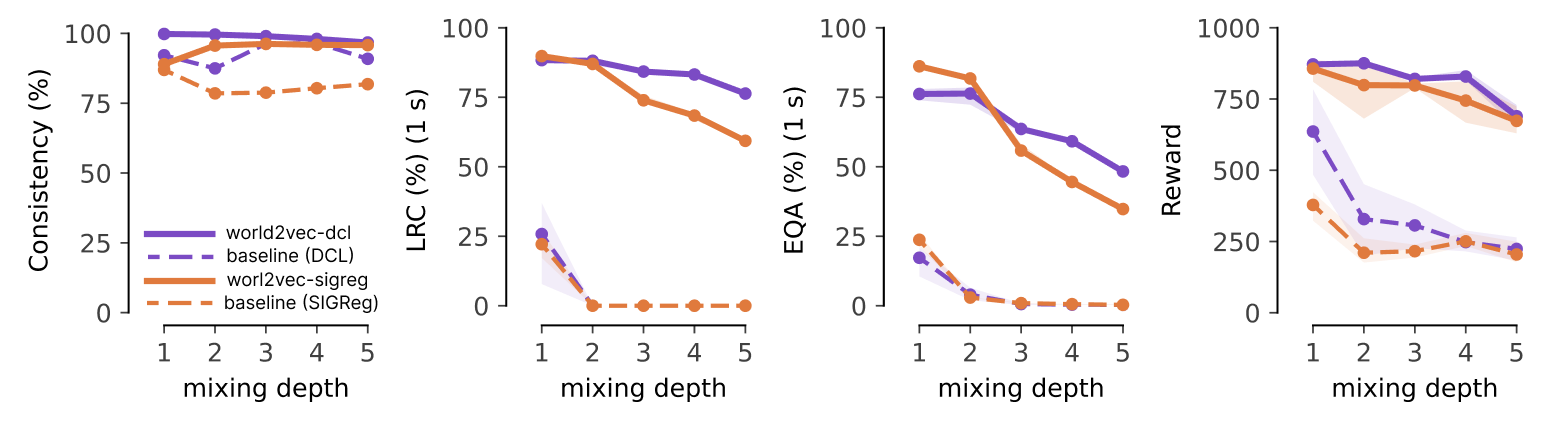}  
    \caption{
        Comparison of \wv{} across increasingly non-linear representations of state of CartPole. The number of mixing layers increases the difficulty of the representation learning task due to an increasingly non-linear transformation of the state.
    }
    \label{fig:mixing}
    \end{figure}

    Model predictive control with \wv{} solves the task in the Reacher and Pendulum environments (\cref{tab:main}). \Wvdcl{} outperforms DreamerV3 on both and TD-MPC2 on Reacher, and improves over LeWM by more than 800 points on each task. \Wvsigreg{} also solves both tasks with slightly lower reward.
    Due to the low reward of LeWM in this evaluation setup, we also benchmark its goal-reaching protocol on Reacher, where \wv{} is on par for goals 25 steps ahead and clearly outperforms LeWM on longer term planning over 200 steps, while planning is 3-8 times faster (\cref{tab:reacher-timing}).
    The CartPole environment remains an open challenge for reward-free world models, although we later show that it is in principle solvable from nonlinear observations (\cref{fig:mixing}).

    To the best of our knowledge, this makes \wv{} the first reward-free, task-agnostic world model that jointly estimates representations and dynamics matching the environment and is competitive at challenging control tasks from pixels while being substantially more efficient.

\paragraph{A sufficiently large training horizon and latent dynamics matching are essential for latent world modeling.}
    We next investigated the importance of the architectural innovations in \wv{}.
    Compared to our baseline, \wv{} uses a multi-step rollout as well as an ODE-formulation as its predictor.
    First, multi-step prediction is a crucial component across all environments.
    Increasing the training horizon drastically improves the model's performance for both \wvdcl{} (\cref{tab:dynamics-dcl}) and \wvsigreg{} (\cref{tab:dynamics-sigreg}).
    For \wvsigreg{}, increasing the horizon from 1 step to 20 steps improves the LRC from $59.3\%$ to $90.2\%$ for pendulum, from $61.5\%$ to $98.7\%$ for reacher, and from $0.0\%$ to $86.4\%$ for cartpole. The model is not sensitive to a particular hyperparameter choice, as long as we move well beyond single step prediction.
\begin{table}[b]\centering\small
    \vspace{-0.5em}
    \caption{%
        Training horizon in steps $(K)$ and in seconds for \wv{} with Euler ODE dynamics; latent width (and SIGReg weight for \wvsigreg{}) selected per cell on validation EQA. All metrics in \%; mean $\pm$ bootstrap standard error over three seeds.
        Bold rows are the configurations used in \cref{tab:main} (\cref{tab:configurations-won}).
        LRC and Reacher EQA differ from there because the ablations use a different evaluation protocol used for model development and selection, outlined in \cref{app:metrics}. 
    }
    \label{tab:dynamics}
    \begin{subtable}[t]{0.49\linewidth}\centering\scriptsize 
    \setlength{\tabcolsep}{1.5pt}
    \caption{\wvdcl{}}\label{tab:dynamics-dcl}
    \begin{adjustbox}{max width=\linewidth}
\begin{tabular*}{\linewidth}{@{\extracolsep{\fill}}lcrP{2.8}P{5.4}P{1.8}@{}}
\toprule
Environment & $K$ (s) & Cons. & LRC$_{1\,\mathrm{s}}$ & EQA$_{1\,\mathrm{s}}$ & State $R^2$ \\
\midrule
pendulum & 1 (0.04 s) & 99.9 & \meanstd{85.7}{1.2} & \meanstd{37.2}{3.3} & \meanstd{97.4}{0.1} \\
 & 10 (0.4 s) & 100.0 & \meanstd{95.3}{0.1} & \meanstd{84.6}{0.2} & \meanstd{98.0}{0.1} \\
 & \won{20 (0.8 s)} & \won{99.7} & \won{\meanstd{93.6}{0.9}} & \won{\meanstd{87.7}{0.6}} & \won{\meanstd{97.7}{0.2}} \\
 & 40 (1.6 s) & 99.1 & \meanstd{92.1}{0.3} & \meanstd{85.3}{0.3} & \meanstd{97.0}{0.1} \\
\midrule
reacher & 1 (0.12 s) & 96.0 & \meanstd{90.1}{2.8} & \meanstd{16.0}{5.4} & \meanstd{70.0}{1.8} \\
 & \won{10 (1.2 s)} & \won{98.2} & \won{\meanstd{97.8}{0.0}} & \won{\meanstd{98.6}{0.0}} & \won{\meanstd{96.3}{0.3}} \\
 & 20 (2.4 s) & 97.3 & \meanstd{97.6}{0.1} & \meanstd{98.2}{0.1} & \meanstd{94.6}{0.4} \\
 & 40 (4.8 s) & 82.1 & \meanstd{98.0}{0.0} & \meanstd{95.8}{0.5} & \meanstd{91.9}{1.2} \\
\midrule
cartpole & 1 (0.02 s) & 95.7 & \meanonly{0.0} & \meanstd{0.6}{0.0} & \meanstd{64.4}{0.5} \\
 & 10 (0.2 s) & 95.3 & \meanstd{89.8}{0.1} & \meanstd{18.2}{0.1} & \meanstd{67.7}{0.2} \\
 & 20 (0.4 s) & 98.5 & \meanstd{93.2}{0.2} & \meanstd{29.3}{0.5} & \meanstd{86.6}{0.2} \\
 & \won{40 (0.8 s)} & \won{93.2} & \won{\meanstd{95.2}{0.1}} & \won{\meanstd{38.7}{0.3}} & \won{\meanstd{79.8}{0.8}} \\
\bottomrule
\end{tabular*}
\end{adjustbox}
 
    \end{subtable}\hfill
    \begin{subtable}[t]{0.49\linewidth}\centering\scriptsize
    \setlength{\tabcolsep}{1.5pt}
    \caption{\wvsigreg{}}\label{tab:dynamics-sigreg}
    \begin{adjustbox}{max width=\linewidth}
\begin{tabular*}{\linewidth}{@{\extracolsep{\fill}}lcrP{3.7}P{0.2}P{1.8}@{}}
\toprule
Environment & $K$ (s) & Cons. & LRC$_{1\,\mathrm{s}}$ & EQA$_{1\,\mathrm{s}}$ & State $R^2$ \\
\midrule
pendulum & 1 (0.04 s) & 99.7 & \meanstd{59.3}{3.7} & \meanstd{32.5}{0.1} & \meanstd{98.3}{0.1} \\
 & 10 (0.4 s) & 92.3 & \meanstd{84.3}{1.1} & \meanstd{77.6}{0.7} & \meanstd{98.3}{0.0} \\
 & \won{20 (0.8 s)} & \won{95.4} & \won{\meanstd{90.2}{0.7}} & \won{\meanstd{85.5}{0.5}} & \won{\meanstd{98.3}{0.0}} \\
 & 40 (1.6 s) & 96.2 & \meanstd{92.8}{0.2} & \meanstd{84.6}{0.8} & \meanstd{98.0}{0.0} \\
\midrule
reacher & 1 (0.12 s) & 93.7 & \meanstd{61.5}{0.2} & \meanstd{24.3}{0.2} & \meanstd{74.5}{0.4} \\
 & 10 (1.2 s) & 92.1 & \meanstd{98.2}{0.2} & \meanstd{97.9}{0.2} & \meanstd{91.5}{0.4} \\
 & \won{20 (2.4 s)} & \won{96.9} & \won{\meanstd{98.7}{0.0}} & \won{\meanstd{98.1}{0.1}} & \won{\meanstd{94.7}{0.2}} \\
 & 40 (4.8 s) & 94.9 & \meanstd{98.5}{0.0} & \meanstd{97.9}{0.0} & \meanstd{94.0}{0.1} \\
\midrule
cartpole & 1 (0.02 s) & 95.5 & \meanonly{0.0} & \meanstd{3.1}{0.2} & \meanstd{38.4}{0.0} \\
 & 10 (0.2 s) & 80.5 & \meanstd{83.5}{0.8} & \meanstd{20.3}{0.6} & \meanstd{38.2}{0.0} \\
 & 20 (0.4 s) & 71.1 & \meanstd{86.4}{0.4} & \meanstd{28.3}{0.4} & \meanstd{38.3}{0.1} \\
 & \won{40 (0.8 s)} & \won{97.2} & \won{\meanstd{88.9}{0.2}} & \won{\meanstd{37.0}{0.3}} & \won{\meanstd{38.9}{0.0}} \\
\bottomrule
\end{tabular*}
\end{adjustbox}
 
    \end{subtable}
\end{table}
\begin{table}[t]\centering\small
    \caption{
    Discrete-time vs.\ continuous-time dynamics. \WV{} trained with $K=20$ rollout steps and a three-frame
    encoder, with discrete MLP transitions ($\rvPhi(\vz,\va)=\mathrm{MLP}(\vz,\va)$) or Euler-integrated ODE
    dynamics, at latent width $d\in\{\text{state dim},32,128\}$ (state dim matches each environment's physical state size). Per environment, the number of negatives (DCL) or the
    SIGReg weight with the best validation EQA\ is kept. State $R^2$ is MLP state decoding; all metrics in \%.
    Each metric is averaged over pendulum, reacher, and cartpole per model seed,
    then mean $\pm$ bootstrap standard error over three seeds. Evaluation protocol as in \cref{tab:dynamics} (\cref{app:metrics}).
    }
    \label{tab:mlp-euler}
    \begin{subtable}[t]{0.49\linewidth}\centering\scriptsize
    \setlength{\tabcolsep}{1.5pt}
    \caption{\wvdcl{}}
    \begin{adjustbox}{max width=\linewidth}
\begin{tabular*}{\linewidth}{@{\extracolsep{\fill}}llrP{1.7}P{1.3}P{2.2}@{}}
\toprule
Dynamics & $d$ & Cons. & LRC$_{1\,\mathrm{s}}$ & EQA$_{1\,\mathrm{s}}$ & State $R^2$ \\
\midrule
Discrete-time & state dim & 88.2 & \meanstd{95.1}{0.1} & \meanstd{67.6}{0.5} & \meanstd{70.5}{0.5} \\
 & 32 & 37.3 & \meanstd{50.9}{1.7} & \meanstd{10.6}{1.3} & \meanstd{39.0}{2.2} \\
 & 128 & 2.2 & \meanstd{0.8}{0.2} & \meanstd{0.3}{0.0} & \meanstd{9.7}{1.5} \\
\midrule
Continuous-time & state dim & 90.0 & \meanstd{95.7}{0.1} & \meanstd{69.2}{0.5} & \meanstd{83.8}{0.4} \\
 & 32 & 98.5 & \meanstd{94.3}{0.1} & \meanstd{71.1}{0.3} & \meanstd{92.8}{0.3} \\
 & 128 & 97.3 & \meanstd{94.0}{0.3} & \meanstd{71.7}{0.2} & \meanstd{93.0}{0.0} \\
\bottomrule
\end{tabular*}
\end{adjustbox}

    \end{subtable}\hfill
    \begin{subtable}[t]{0.49\linewidth}\centering\scriptsize
    \setlength{\tabcolsep}{1.5pt}
    \caption{\wvsigreg{}}
    \begin{adjustbox}{max width=\linewidth}
\begin{tabular*}{\linewidth}{@{\extracolsep{\fill}}llrP{7.1}P{7.5}P{1.8}@{}}
\toprule
Dynamics & $d$ & Cons. & LRC$_{1\,\mathrm{s}}$ & EQA$_{1\,\mathrm{s}}$ & State $R^2$ \\
\midrule
Discrete-time & state dim & 63.3 & \meanstd{93.7}{0.1} & \meanstd{62.3}{0.9} & \meanstd{62.3}{0.8} \\
 & 32 & 95.1 & \meanstd{90.0}{0.7} & \meanstd{65.2}{0.9} & \meanstd{64.8}{0.2} \\
 & 128 & 75.2 & \meanstd{60.1}{2.8} & \meanstd{28.7}{1.5} & \meanstd{69.0}{1.4} \\
\midrule
Continuous-time & state dim & 82.4 & \meanstd{92.5}{0.2} & \meanstd{64.2}{0.9} & \meanstd{69.9}{1.8} \\
 & 32 & 97.3 & \meanstd{92.4}{0.3} & \meanstd{70.4}{0.2} & \meanstd{77.1}{0.1} \\
 & 128 & 95.8 & \meanstd{76.9}{7.1} & \meanstd{42.3}{7.5} & \meanstd{72.2}{0.6} \\
\bottomrule
\end{tabular*}
\end{adjustbox}
 
    \end{subtable}
\end{table}
    Second, we test the importance of the Neural ODE dynamics model against a discrete-time parameterization in which we use $\rvPhi(\vz,\va)=\mathrm{MLP}(\vz,\va)$ (\cref{sec:setup}).
    Across both \wvsigreg{} and \wvdcl{} the Euler ODE formulation clearly improves over the discrete MLP dynamics (\cref{tab:mlp-euler}).
    In the regimes of lower-dimensional latent spaces, MLP dynamics can still be used to learn dynamics that outperform the SOTA baselines in terms of LRC, achieving LRC of $95.1\%$ on \wvdcl{} at state dimension and $90.0\%$ on \wvsigreg{} at $d=32$ with discrete-time dynamics.
    The Euler ODE formulation results in higher Consistency scores, higher (or on par) LRC, higher EQA, and better state decoding when comparing the best-performing models across latent dimensionalities.
    This is in line with the analysis by \citet{laiz2026understanding}, who show that latent world models with standard predictors fail to model velocity information. Modeling latent state change $\dot \vz$ vs. state $\vz$ itself counteracts this failure.

\begin{wrapfigure}{r}{0.52\textwidth}
    \vspace{-1.4em}
    \centering
    \includegraphics[width=\linewidth]{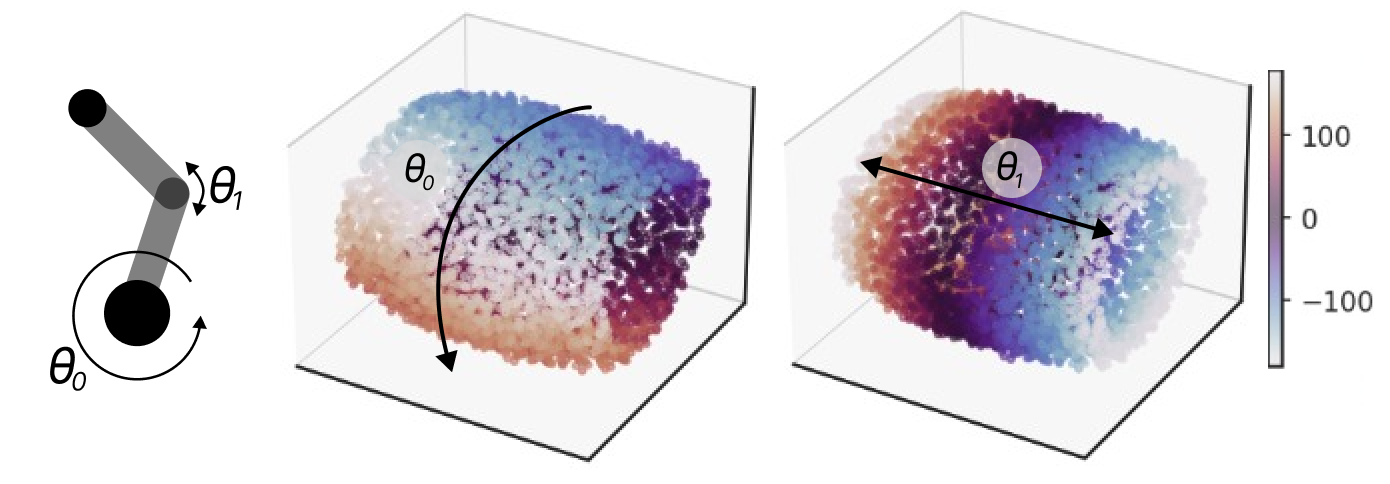}
    \caption{PCA of Reacher embeddings, colored by
        state variables $\theta_0$ (left, circular) and $\theta_1$ (right, linear).}
    \label{fig:topology}
    \vspace{-1.5em}
    \end{wrapfigure}

\paragraph{Embedding geometry qualitatively reflects environment structure.}
    As a qualitative assessment of the coordinates learned by \wv{}, we visualized a PCA projection of the embeddings obtained on the Reacher environment in \cref{fig:topology}. This environment can be represented
    through two angles $(\theta_0, \theta_1)$ that describe the current configuration of the two arms. 
    While $\theta_0$ is executing full 360 degree movements during data collection, $\theta_1$ is constrained to rotating left and right relative to the first arm without observing full circular movements. This is reflected in the embedding, where $\theta_0$ is encoded with circular boundaries, while $\theta_1$ is represented linearly with open boundaries.

\begin{figure}[t]\centering
        \includegraphics[width=\linewidth]{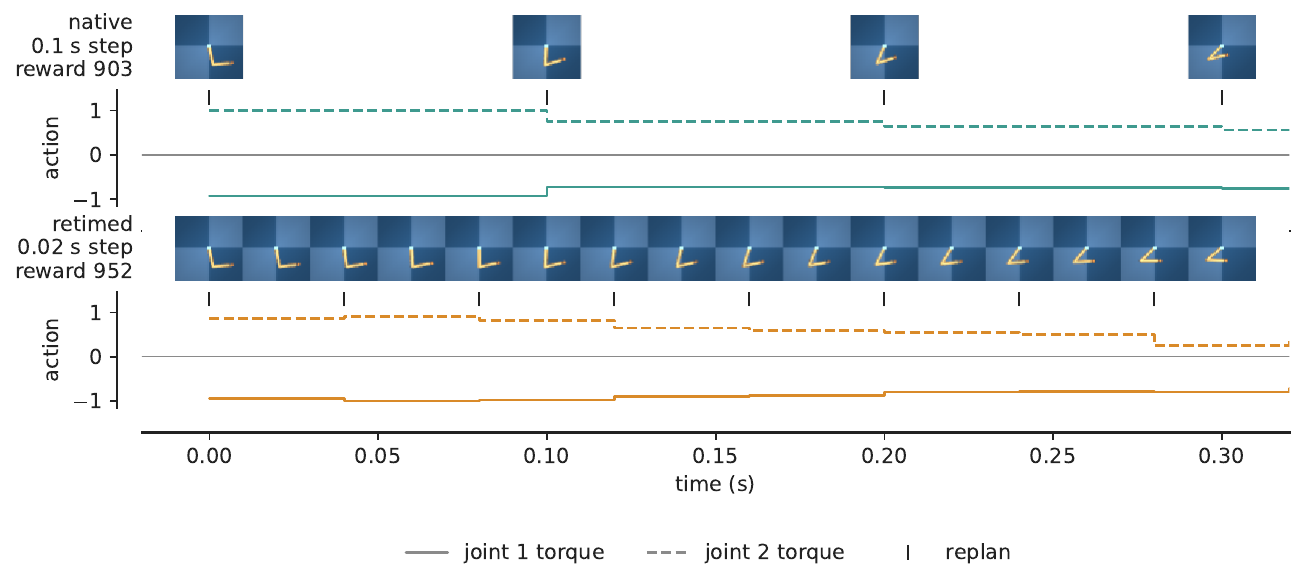}
        \caption{%
            Control across time scales for Reacher.
            The same \wvdcl{} model (K=20), start state and target, planned at the 10\,Hz rate it was trained on (top, \emph{native}) and retimed to 50\,Hz by reducing the step size (bottom, \emph{retimed}). 
            Frames are the planner's observations, drawn at the time they were taken; below each strip, the two joint torques applied and a tick at every replanning (10\,Hz native, 25\,Hz retimed). Episode returns: 903 (native) and 952 (retimed).
        }
        \label{fig:reacher_retime}
    \end{figure}

\paragraph{Continuous-time ODE parameterization allows flexible sampling.}
    An important benefit of modeling state change through ODE dynamics is the possibility of adapting to a different sampling rate at inference time. %
    As a demonstration, we train a \wv{} model on Reacher at 10\,Hz and use it for model predictive control at this rate, and at 50\,Hz by reducing the step size of the Euler integrator accordingly (\cref{fig:reacher_retime}). Holding each action for two time steps gives a control rate of 25\,Hz. The retimed controller performs at least as well as the native one (returns of 952 and 903).

    \paragraph{Empirical identifiability of \wv{}.}
        Identifiable representation learning denotes the property where two models converge to the same representation.
        This property has been studied extensively in the context of representation learning and world models, and for the given objective, we expect  that the representations derived from training will be consistent up to an affine transformation, and that dynamics will be modeled faithfully over time\citep{zimmermann_contrastive_2021,mikulasch2026understanding,mikulasch2026predictive} as detailed in \cref{sec:towards-mechanistic-world-models};
        note that we do not expect to recover the state linearly due to missing variability\citep{roeder2021linear} along some dimensions in the considered (noiseless) systems.
        We focus our analysis on the CartPole environment, which was not solvable by any reward-free world model from raw observations. We study this environment in the setting from \citet{laiz2026understanding}, where we replace pixel information by an increasingly non-linear mixing function, as detailed in the experimental setup.
        Under such a systematic increase in task difficulty, we observe a decline across the different metrics.
        LRC and EQA decline as the mixing depth increases, from about 90\% and 75--85\% at one mixing layer to 60--75\% and 35--50\% at five layers, while the one-step baselines collapse for two or more layers. \Wvdcl{} degrades least, keeping an LRC above 75\% at five layers (\cref{fig:mixing}).
        Reward declines only slightly, from about 850 to 700, so both \wv{} variants still solve CartPole from non-linearly mixed observations, whereas the baselines remain around 200--300.
        The gap between EQA in this task and under pixel observations is modest; we anticipate the possibility to transfer this result to pixel observations with additional improvements in the dynamics model estimation in future work.

\section{Symbolic regression for understanding latent mechanisms}\label{sec:sindy}
    
        \begin{figure}[t]
            \centering
            \includegraphics[width=\linewidth]{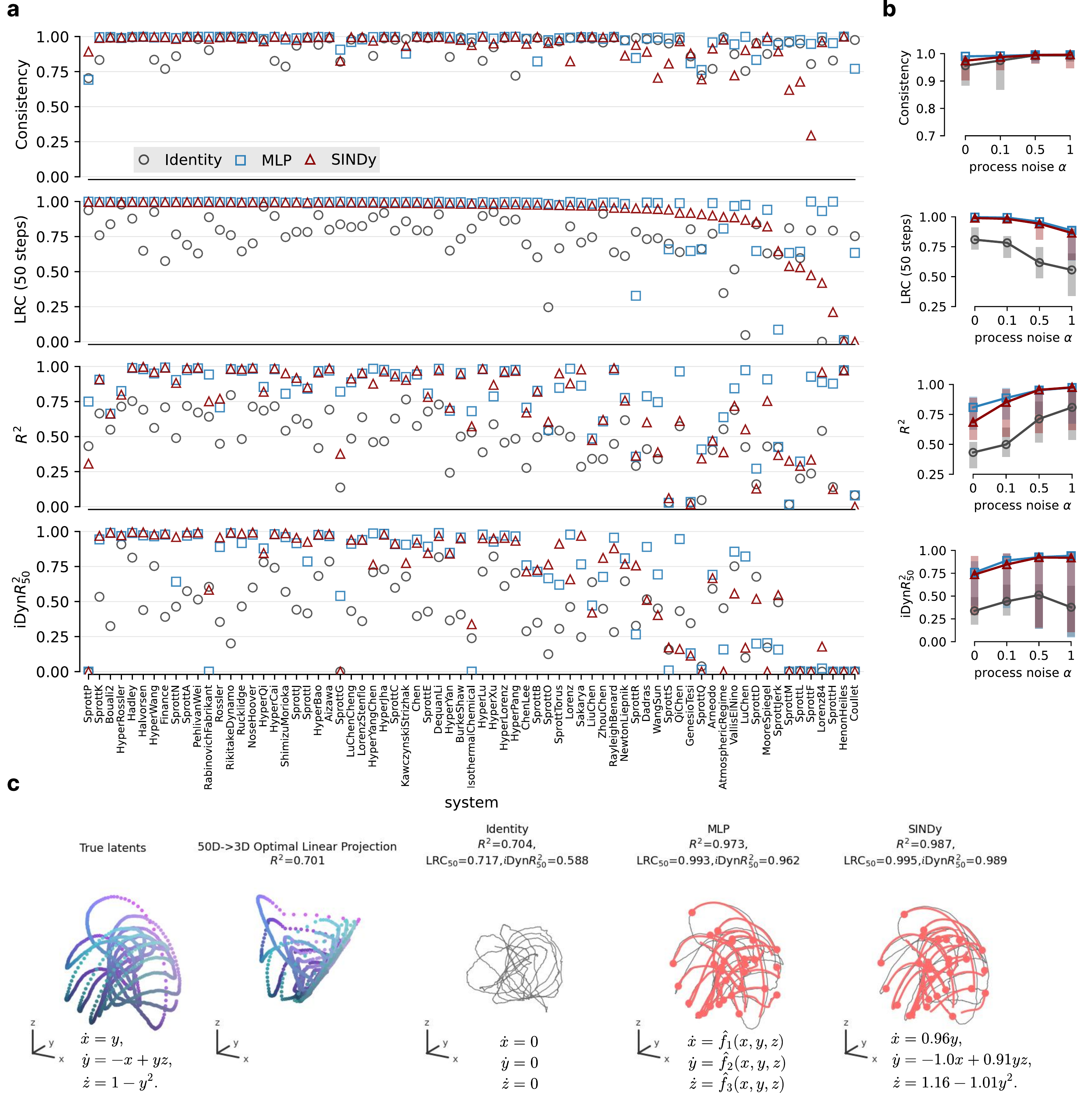}
            \caption{%
                Latent dynamics recovery, representation consistency, and affine state recovery across 71 \texttt{dysts} systems.
                    \textbf{(a)} Consistency of ensembles is displayed on top, followed by $\mathrm{LRC}_{50}$, affine recovery $R^2$, and iDyn$R^2_{50}$. $\mathrm{LRC}_{50}$ and iDyn$R^2_{50}$ compare 50-step predictions with encoded futures and true rollouts after affine alignment, respectively.
                    The points show means over three seeds for 71 \texttt{dysts} systems at process-noise level $\alpha=0.1$, ordered by decreasing $\mathrm{LRC}_{50}$ of SINDy models. Identity, MLP and SINDy dynamics models are shown as grey circles, blue squares and red triangles, respectively.
                    \textbf{(b)} Lines and shaded bars show medians and 25th--75th percentiles across systems for $\alpha\in\{0,0.1,0.5,1.0\}$ for consistency, $\mathrm{LRC}_{50}$, $R^2$, and iDyn$R^2_{50}$.
                    \textbf{(c)} For the selected system, SprottA, true states and their optimal linear reconstruction from 50-dimensional observations share corresponding point colours; grey curves show learned embeddings and red curves show predicted rollouts, with dots marking initial states, all in ground-truth coordinates.
                    For each dynamics model, as well as the true system, the equations governing the dynamics are displayed in the bottom of the rollout figures.
            }
            \label{fig:sindy}
        \end{figure}

    \subsection{Experimental setup}
    To allow \wv{} to discover interpretable latent dynamics, we integrate our framework with SINDy \citep{sindy}. The resulting \wvsindy{} algorithm represents the flow field of a dynamical system with the action-free form of \cref{eq:library}, $\dot{\vz}=\boldsymbol{\Xi}\,\boldsymbol{\theta}(\vz)\in\mathbb{R}^{d}$, where $\boldsymbol{\theta}(\vz)\in\mathbb{R}^{p}$ stacks all monomials of the latent coordinates up to degree $q$, including the constant, and the $i$-th row of the sparse coefficient matrix $\boldsymbol{\Xi}\in\mathbb{R}^{d\times p}$ defines the governing equation of $z_i$.
    We compare this dynamics model to identity dynamics as a trivial baseline, and to an MLP dynamics model as in the previous experiments.
    All three models are evaluated on 71 strange chaotic attractors from the \texttt{dysts} database \citep{dysts}, a benchmark previously used to evaluate SINDy \citep{sindy-dysts}.
    We extend this benchmark by mixing the physical state variables using the same methodology as in the previous section, and add process noise with standard deviation scale $\alpha$.
    We train with one-step prediction ($K=1$) using RK4 integration rather than the Euler rollout of \cref{eq:euler}.
    Additional improvements with higher values for $K$ are expected, but left to future extensions of this work. Full details of the experimental setup are provided in \cref{app:sindy}.

\subsection{Results}
        
    \paragraph{\Wvsindy{} outperforms the identity baseline and matches an MLP vector field.}
        Across the latent state and dynamics recovery metrics, \wvsindy{} substantially improves over the identity predictor and performs comparably to the MLP model (\cref{fig:sindy}a). For latent process noise $\alpha = 0.1$, we recover highly consistent ensembles across all three dynamics model families.
        For SINDy and MLP, this consistency is accompanied by high affine recovery $R^2$ for many of the 71 systems. Furthermore, for multiple systems, the SINDy and MLP rollouts agree with both the encoder's future embeddings and the aligned ground-truth rollouts, as measured by $\mathrm{LRC}_{50}$ and iDyn$R^2_{50}$ (which measures the recovery of the dynamical system itself \citep{dcl}).
        
    \paragraph{Increasing process noise enables linear state recovery.}

        Adding process noise to the ground truth states adds variations across all dimensions, which should enable linear recovery of the state information\citep{dcl,mikulasch2026understanding}.
        We empirically investigate this property across four noise levels, $\alpha \in \{0,0.1,0.5,1.0\}$, showing medians and interquartile ranges (25th--75th percentiles) across systems for all four metrics (\cref{fig:sindy}b). As $\alpha$ increases from $0$ to $1.0$, median affine recovery $R^2$ rises from approximately $0.8$ to $0.98$ for MLP, from $0.7$ to $0.97$ for SINDy, and from $0.4$ to $0.8$ for the identity model which lines up with previous findings on latent dynamics learning\citep{dcl}.
        Consistency is high across all models, but slightly increases until it reaches approximately $0.99$ for all three models at $\alpha=1.0$.
        The process noise sets an upper limit on the LRC (optimal latent prediction is no longer possible), which decreases as noise increases.
        In cases where linear state discovery is obtained ($\alpha > 0.5$) the median iDyn$R^2_{50}$ is a meaningful metric to measure the faithfulness of the dynamics in absence of LRC; indeed it approaches close to 100\% for high noise levels, and substantially outperforms the identity baseline.
        
    \paragraph{\Wvsindy{} can recover analytical equations of latent dynamics.}
        Most relevant for mechanistic interpretability, unlike the ``black-box'' flow field of the MLP, the SINDy parameterization enables recovery of interpretable equations from the nonlinearly mixed high-dimensional observations.
        Among the ten systems with the highest validation $\mathrm{LRC}$ for SINDy, we visualize the system with the smallest mean pairwise Frobenius distance between aligned coefficient matrices.
        To express the recovered equations in the ground-truth coordinate basis, we estimate the affine indeterminacy via least-squares regression from the learned embeddings to the ground-truth states, transform the SINDy coefficient matrix accordingly, threshold coefficients at 0.1, and visualize the resulting equation (\cref{fig:sindy}c).
        In future work, it would be interesting to extend these experiments with the unsupervised symbolic regression technique by \citet{dysco} and explore options to replace the fixed library with end-to-end learning of basis functions using tools from mechanistic interpretability\citep{cunningham2023sparse,choi2026beyond}.

\section{Conclusion}\label{sec:conclusion}

    We introduced \wv{}, a framework for learning mechanistic world models by jointly fitting representations and latent dynamics from observations.
    To the best of our knowledge, \wv{} is the first reconstruction- and reward-free algorithm to jointly learn explicit dynamics and achieve successful control in environments with autonomous dynamics.
    This is enabled by combining multi-step rollouts, latent dynamics matching, and model selection using metrics grounded in identifiability theory.
    Despite these advances, the environments considered remain relatively simple. We see improving and scaling data collection (potentially through online data collection), training across multiple environments, and studying the identifiability of our objective functions under less idealized conditions as promising directions for extending \wv{} towards solving more complex tasks.

\subsection*{Author Contributions}
    StS and TS conceptualized the method. TS developed the original \wv{} implementation, trained the initial models, and contributed to the formal analysis. RGL designed the benchmarking setup, implemented the baselines and contributed to the formal analysis. TS and KS conducted and analyzed the symbolic regression experiments. All authors contributed to writing and revising the manuscript.

\subsection*{Acknowledgements}
    This work was supported by the Helmholtz Association's Initiative and Networking Fund (IVF) through a Helmholtz AI grant, a Helmholtz Imaging grant under the UNLOCK benchmarking initiative, and computational resources via the HAICORE@KIT and HAICORE@FZJ partitions.

\subsection*{AI use statement}
    We used generative AI tools to assist with drafting and editing the manuscript, implementing and running experimental code, and creating plotting code for the figures.
    All AI-assisted text and code were reviewed by the authors. Experimental outputs and figures were checked for correctness, and the authors take full responsibility for the final content of this work.

\bibliographystyle{unsrtnat}
\bibliography{references}

\clearpage
\appendix
\supplementaryfloats

\section{Training objectives}\label{app:objectives}

All \wv{} models are trained via gradient descent to minimize \cref{eq:rollout}. By default, we set the optional weighting factor $w_j$ for the multi-step dynamics matching described in \cref{eq:rollout} to be uniform and normalized to $w_j = 1/K$ for a $K$-step rollout.

\Wvsigreg{} and \wvdcl{} differ in their training objectives in the way they regularize the encoder model. Intuitively, choosing either one of these methods specifies a certain type of prior on how the embedding space is expected to be distributed. By choosing \wvsigreg{}, one specifies a Gaussian prior on the embedding space\cite{lejepa}, whereas choosing \wvdcl{} corresponds to specifying a Uniform distribution over the support of the dataset\citep{zimmermann_contrastive_2021,contrastive_noise}. Other marginal distributions are possible\citep{dcl} but not further considered here. Therefore, the choice of SIGReg vs. DCL is mainly a choice about which prior is better suited for the dataset/environment at hand. 

\paragraph{\Wvsigreg{}} uses the SIGReg regularizer as in LeJEPA \cite{lejepa}, which pushes the distribution of embeddings towards an isotropic Gaussian $\mathcal N(\vzero,\mI)$. For a set of embeddings $\{\vz_b\}_{b=1}^B$, SIGReg projects them onto a set $\mathcal A$ of directions $\vu$ drawn uniformly on the unit sphere and compares each one-dimensional projection with $\mathcal N(0,1)$ using the Epps--Pulley test statistic,
\begin{equation}
    \mathrm{SIGReg}\big(\{\vz_b\}_{b=1}^B\big)=\frac{1}{|\mathcal A|}\sum_{\vu\in\mathcal A}B\int_{-\infty}^{\infty}w(t)\,\big|\hat\varphi_{\vu}(t)-e^{-t^2/2}\big|^2\,dt,
    \qquad
    \hat\varphi_{\vu}(t)=\frac1B\sum_{b=1}^{B}e^{\,\mathrm{i}t\,\vu^\top\vz_b},
\end{equation}
where $\hat\varphi_{\vu}$ is the empirical characteristic function of the projections and $e^{-t^2/2}$ is the characteristic function of $\mathcal N(0,1)$. By the Cram\'er--Wold theorem, matching all one-dimensional projections matches the joint distribution.
As in LeWM \cite{lewm}, SIGReg is applied to a batch of embeddings. With multi-step rollouts, each starting point contributes an embedding at every step $j=0,\dots,K$: the embeddings at the starting state and at the $K$ prediction targets. We apply SIGReg independently to the batch of embeddings at each step and average the results:
\begin{equation}
    \mathcal R_{\text{SIG}}(\rvh)=\frac{1}{K+1}\sum_{j=0}^{K}\mathrm{SIGReg}\big(\{\rvh(\mX_{n_b+j})\}_{b=1}^{B}\big),
\end{equation}
where $n_1,\dots,n_B$ are the starting points of a batch. The encodings at one step come from independent starting points, whereas encodings along one rollout are strongly correlated.
As in the released implementations of LeJEPA and LeWM, we use the weight $w(t)=e^{-t^2/2}$ and evaluate the integral over $[-3,3]$ with the trapezoidal rule: since the integrand is even in $t$, it suffices to evaluate it at $17$ equally spaced nodes on $[0,3]$ and double the result. We use $|\mathcal A|=1024$ directions ($64$ for nonlinear-mixing observations), resampled at every iteration.
We select $\lambda\in\{0.01,0.1,1\}$ for pixel observations and $\lambda\in\{0.001,0.01\}$ for nonlinear-mixing observations on validation EQA.

\paragraph{\Wvdcl{}} is based on a variant of the contrastive regularizer used in DCL \cite{dcl}. We additionally sample $M$ negative samples $m_1,\dots,m_M\sim p_{\mathcal D}$ uniformly across the training set (in practice, one set per batch, shared by all starting points) and measure how close the encoded starting state is to them,
\begin{equation}
    S_n=\sum_{i=1}^{M}e^{-\|\rvh(\mX_n)-\rvh(\mX_{m_i})\|_2^2}.
\end{equation}
With the prediction error $d_{n,j}=\|\hat\vz_{n,j}-\rvh(\mX_{n+j})\|_2^2$ from \cref{eq:rollout}, the regularizer is
\begin{equation}\label{eq:app-dcl}
\begin{split}
    \mathcal R_{\text{DCL}}(\rvh,\hat\rvf)
    &=\E_{n,\,m_{1:M}}\!\left[\sum_{j=1}^K w_j\log\!\big(e^{-d_{n,j}}+S_n\big)\right]\\
    &=\underbrace{\E_{n,\,m_{1:M}}\big[\log S_n\big]}_{\text{uniformity (encoder only)}}
    +\underbrace{\E_{n,\,m_{1:M}}\!\left[\sum_{j=1}^K w_j\log\!\Big(1+\frac{e^{-d_{n,j}}}{S_n}\Big)\right]}_{\geq 0},
\end{split}
\end{equation}
where the second line uses $\sum_j w_j=1$.
The first part depends only on the encoder and is the core of the regularizer. It is the uniformity term of contrastive learning \cite{wang2020understanding}: it penalizes the encoded state for being close to the encoded negatives, most strongly to the nearest ones, thereby spreading the embeddings of different observations across the embedding space. This is how \wvdcl{} prevents collapse.
In contrast to DCL, where the negatives are compared with the prediction of the dynamics model, we compare them with the encoded starting state. The uniformity term contributes no gradient to the dynamics model, and the prior towards a uniform embedding acts on the encoder alone.
The second part compares each prediction error with the distances to the negatives. It keeps the objective bounded below, since each horizon contributes $d_{n,j}+\log(e^{-d_{n,j}}+S_n)\geq 0$ to \cref{eq:rollout}; without it, scaling up the embedding could decrease the objective without bound.
It is the only place where the dynamics model enters the regularizer, and it only rescales the gradient of the prediction term: the derivative of each horizon's contribution with respect to $d_{n,j}$ is $1-e^{-d_{n,j}}/(e^{-d_{n,j}}+S_n)\in(0,1)$. The dynamics model is thus always trained to reduce its prediction error.
We use $M\in\{128,512,1024\}$ for pixel observations and $M=1024$ for nonlinear-mixing observations.

\section{Metrics}\label{app:metrics}

    \paragraph{Latent Rollout Consistency (LRC).} Following \citet{laiz2026understanding}, LRC measures the agreement of the encoder and dynamics models in an unsupervised way.
    Given a sequence of observations $\{\vx_t\}^T_{t=0}$, we encode this sequence into latent space using the encoder $\vz_t=\rvh(\vx_t)$. The latent dynamics model predicts $\hat{\vz}_{t+k}= \rvPhi^{k}(\vz_{t})$ and we compute the LRC for horizon $k$ as the score between $\vz_{t+k}$ and $\hat{\vz}_{t+k}$.
    
    \paragraph{Evaluation pipelines.} \Cref{tab:main,tab:configurations-won,tab:lewm-protocol} use a separate evaluation pipeline shared with all baselines; the ablations in \cref{tab:dynamics,tab:mlp-euler} use our world2vec training pipeline. The latter averages the LRC $R^2$ uniformly over latent dimensions instead of weighting by variance, and evaluates the 8.3\,Hz Reacher models on the 8.3\,Hz test split instead of every third frame of the 25\,Hz test split with averaged actions. LRC, and EQA for Reacher at 8.3\,Hz, should therefore be compared within a table.

    \paragraph{Equivariance Accuracy (EQA)}. Following \citet{schmidt2025ebc}, EQA is a retrieval-based extension of LRC. It measures how well we can recover a future observation via KNN decoding in an unsupervised way.
    We encode a sequence of observations; we predict the future latents $\hat{\vz}_{t+k}$ but then we do a KNN lookup to identify the embedding in latent space that is closest to $\hat{\vz}_{t+k}$. To be able to do this, we randomly sample $N=512$ observations from the training dataset, we add the observations $\{\vx_t\}^T_{t=0}$ we compute our metric on to the pool of candidates. We encode these into latent space, find the closest latent, compute a Top-k accuracy score on this classification task, and report Top-1 accuracy.
    
    \paragraph{Embedding Consistency (Cons.)} measures how well multiple models (different seeds) trained on the same data agree with each other.
    This is done by taking the same set of observations, encoding them into latent space with each model $\mZ^{(k)} = \rvh_{k}(\mX)$. Then we fit a linear regression to map $\mZ^{(i)} \xrightarrow{} \mZ^{(j)} $ for every $i,j \in [1, ..., K]$, we score how well each of those mappings works via $R^2$ and average these scores to arrive at the embedding consistency score. 
    
    \paragraph{Cumulative Reward} measures how well a model performs in solving a given task. We use the reward as measured by the MuJoCo-based DeepMind control environments: Pendulum (swingup), Reacher (hard), Cartpole (swingup).
    We set episode length and control rate to a) match the setup for LeWM as used in \citet{lewm} and b) constrain per-episode reward to be within 0 and 1000 as maximum cumulative episode reward.
    
    \paragraph{State Decoding} measures how well we can decode the state provided by the environment. We split the validation dataset into two parts: one to train a decoder (Linear Regression or MLP decoder) and the second part on which we evaluate these. 
    
    \paragraph{Identifiable Latents $R^2$ (iLat$R^2$)} measures how well the encoder recovers the true latent representation up to an affine transform. This is the metric as it is proposed and used in DCL \citep{dcl} and corresponds to state decoding via an affine map (linear regression). 
    
    \paragraph{Identifiable Dynamics $R^2$ (iDyn$R^2$)} measures how well the recovered dynamics model aligns with the ground truth latent dynamics models.
    This metric corresponds to the ``Dyn$R^2$'' metric proposed for DCL \citep{dcl} at horizon k. As part of this metric, we fit an affine map from the encoder's embeddings into the ground truth latent space. The same map as the Identifiable Latents $R^2$ metric uses. However, instead of scoring the embeddings, we score how well future embeddings align with the ground truth dynamics.

    \begin{table}[ht]\centering\small
        \caption{%
            Configurations used in \cref{tab:main}, per environment.
            $d$ is the latent width and $\lambda$ the regularizer weight.
            $K$ (s) is the training rollout in dataset steps and in seconds.
            Cons.\ is the held-out affine $R^2$; Cons., LRC and EQA in \%.
            Reward is that row's dm\_control return (maximum 1000).
            Mean $\pm$ bootstrap standard error over seeds.
            Both Reacher models are shown: 25\,Hz, and 8.3\,Hz rolled out on every third frame of the 25\,Hz test data.
            Reacher reward averages both IK branches.
            The 8.3\,Hz \wv{} rows are the configurations shown in bold in \cref{tab:dynamics}, scored with the evaluation pipeline (\cref{app:metrics}).
        }
        \label{tab:configurations-won}
        \begin{adjustbox}{max width=\linewidth}
\setlength{\tabcolsep}{3.5pt}
\begin{tabular*}{\linewidth}{@{\extracolsep{\fill}}llcccrP{7.1}P{1.8}P{141}@{}}
\toprule
Method & Environment & $d$ & $\lambda$ & $K$ (s) & Cons. & LRC$_{1\,\mathrm{s}}$ & EQA$_{1\,\mathrm{s}}$ & Reward \\
\midrule
Baseline (SIGReg, $\approx$ LeWM)
 & Pendulum & 32 & 1 & 1 (0.04\,s) & 76.2 & \meanstd{40.9}{4.0} & \meanstd{29.7}{1.7} & \meanstd{12}{10} \\
 & Reacher 25\,Hz & 6 & 0.1 & 1 (0.04\,s) & 37.7 & \meanstd{1.4}{1.2} & \meanstd{1.8}{0.2} & \meanstd{97}{21} \\
 & Reacher 8.3\,Hz & 32 & 1 & 1 (0.12\,s) & 88.8 & \meanstd{71.4}{1.1} & \meanstd{27.3}{1.5} & \meanstd{87}{6} \\
 & CartPole & 32 & 1 & 1 (0.02\,s) & 67.1 & \meanonly{0.0} & \meanstd{3.9}{0.9} & \meanstd{201}{4} \\
\midrule
Baseline (DCL)
 & Pendulum & 128 & -- & 1 (0.04\,s) & 99.5 & \meanstd{69.0}{2.3} & \meanstd{23.9}{0.9} & \meanstd{3}{3} \\
 & Reacher 25\,Hz & 6 & -- & 1 (0.04\,s) & 61.3 & \meanstd{38.4}{7.1} & \meanstd{2.0}{0.4} & \meanstd{56}{14} \\
 & Reacher 8.3\,Hz & 6 & -- & 1 (0.12\,s) & 93.9 & \meanstd{84.7}{0.7} & \meanstd{8.9}{0.8} & \meanstd{104}{8} \\
 & CartPole & 32 & -- & 1 (0.02\,s) & 71.5 & \meanonly{0.0} & \meanstd{0.4}{0.2} & \meanstd{154}{8} \\
\midrule
\wvsigreg{}
 & Pendulum & 32 & 1 & 20 (0.8\,s) & 95.8 & \meanstd{98.5}{0.1} & \meanstd{85.6}{1.4} & \meanstd{434}{141} \\
 & Reacher 25\,Hz & 32 & 1 & 20 (0.8\,s) & 96.3 & \meanstd{99.0}{0.1} & \meanstd{96.3}{0.4} & \meanstd{759}{10} \\
 & Reacher 8.3\,Hz & 32 & 1 & 20 (2.4\,s) & 97.6 & \meanstd{95.6}{0.2} & \meanstd{77.6}{1.8} & \meanstd{570}{20} \\
 & CartPole & 32 & 1 & 40 (0.8\,s) & 97.4 & \meanstd{94.1}{0.1} & \meanstd{40.4}{1.3} & \meanstd{221}{1} \\
\midrule
\wvdcl{}
 & Pendulum & 128 & -- & 20 (0.8\,s) & 99.7 & \meanstd{99.8}{0.0} & \meanstd{87.5}{0.8} & \meanstd{826}{0} \\
 & Reacher 25\,Hz & 32 & -- & 20 (0.8\,s) & 98.5 & \meanstd{99.0}{0.0} & \meanstd{97.3}{0.2} & \meanstd{618}{4} \\
 & Reacher 8.3\,Hz & 128 & -- & 10 (1.2\,s) & 95.1 & \meanstd{95.7}{0.1} & \meanstd{85.4}{1.4} & \meanstd{895}{16} \\
 & CartPole & 128 & -- & 40 (0.8\,s) & 91.8 & \meanstd{97.7}{0.1} & \meanstd{39.0}{0.1} & \meanstd{232}{5} \\
\bottomrule
\end{tabular*}
\end{adjustbox}

    \end{table}

\section{Controllers}\label{app:controllers}

The rewards in \cref{tab:main,tab:configurations-won} use model predictive control with the cross-entropy method (CEM). CEM samples action sequences, keeps the best ones, and refits the sampling distribution. \Cref{tab:controllers} is the planner selected for each model and environment in \cref{tab:configurations-won}.

\begin{table}[ht]
\centering
\small
\caption{%
    CEM planners for the configurations in \cref{tab:configurations-won}.
    $d$ is the latent width and $\lambda$ the SIGReg weight.
    $K$ (s) is the training rollout in dataset steps and in seconds.
    Iters are the CEM updates per plan.
    Lookahead is horizon $\times$ hold, in model steps and in seconds.
    Reward is that planner's dm\_control return (maximum 1000), mean $\pm$ bootstrap standard error over seeds.
}
\label{tab:controllers}
\setlength{\tabcolsep}{3.2pt}
\begin{adjustbox}{max width=\linewidth}
\begin{tabular}{llccccrrrrlP{141}}
\toprule
& & & & & \multicolumn{6}{c}{Planner} & \\
\cmidrule(lr){6-11}
Method & Environment & $d$ & $\lambda$ & $K$ (s) & Iters & Horizon & Hold & Exec. & Lookahead & Cost & Reward \\
\midrule
Baseline (SIGReg, $\approx$ LeWM)
 & Pendulum & 32 & 1 & 1 (0.04\,s) & 6 & 12 & 2 & 1 & 24 (0.96\,s) & Ramp & \meanstd{12}{10} \\
 & Reacher 25\,Hz & 6 & 0.1 & 1 (0.04\,s) & 6 & 10 & 2 & 1 & 20 (0.8\,s) & Flat & \meanstd{97}{21} \\
 & Reacher 8.3\,Hz & 32 & 1 & 1 (0.12\,s) & 6 & 4 & 2 & 1 & 8 (0.96\,s) & Flat & \meanstd{87}{6} \\
 & CartPole & 32 & 1 & 1 (0.02\,s) & 6 & 6 & 4 & 1 & 24 (0.48\,s) & Ramp & \meanstd{201}{4} \\
\midrule
Baseline (DCL)
 & Pendulum & 128 & -- & 1 (0.04\,s) & 6 & 12 & 2 & 1 & 24 (0.96\,s) & Ramp & \meanstd{3}{3} \\
 & Reacher 25\,Hz & 6 & -- & 1 (0.04\,s) & 6 & 5 & 5 & 1 & 25 (1\,s) & Flat & \meanstd{56}{14} \\
 & Reacher 8.3\,Hz & 6 & -- & 1 (0.12\,s) & 6 & 4 & 2 & 1 & 8 (0.96\,s) & Flat & \meanstd{104}{8} \\
 & CartPole & 32 & -- & 1 (0.02\,s) & 6 & 24 & 2 & 1 & 48 (0.96\,s) & Ramp & \meanstd{154}{8} \\
\midrule
\wvsigreg{}
 & Pendulum & 32 & 1 & 20 (0.8\,s) & 6 & 12 & 2 & 1 & 24 (0.96\,s) & Ramp & \meanstd{434}{141} \\
 & Reacher 25\,Hz & 32 & 1 & 20 (0.8\,s) & 12 & 12 & 4 & 2 & 48 (1.92\,s) & Flat & \meanstd{759}{10} \\
 & Reacher 8.3\,Hz & 32 & 1 & 20 (2.4\,s) & 6 & 4 & 2 & 1 & 8 (0.96\,s) & Flat & \meanstd{570}{20} \\
 & CartPole & 32 & 1 & 40 (0.8\,s) & 6 & 6 & 4 & 1 & 24 (0.48\,s) & Ramp & \meanstd{221}{1} \\
\midrule
\wvdcl{}
 & Pendulum & 128 & -- & 20 (0.8\,s) & 6 & 12 & 2 & 1 & 24 (0.96\,s) & Ramp & \meanstd{826}{0} \\
 & Reacher 25\,Hz & 32 & -- & 20 (0.8\,s) & 12 & 12 & 4 & 2 & 48 (1.92\,s) & Flat & \meanstd{618}{4} \\
 & Reacher 8.3\,Hz & 128 & -- & 10 (1.2\,s) & 6 & 4 & 2 & 1 & 8 (0.96\,s) & Flat & \meanstd{895}{16} \\
 & CartPole & 128 & -- & 40 (0.8\,s) & 6 & 12 & 2 & 1 & 24 (0.48\,s) & Ramp & \meanstd{232}{5} \\
\bottomrule
\end{tabular}
\end{adjustbox}
\end{table}

\clearpage
\section{Datasets}\label{app:dataset_collection}

All pixel world models are trained on data from the Pendulum, Reacher, and Cartpole environments of the DeepMind Control Suite \citep{control_suite}, rendered as 64$\times$64 RGB frames (\cref{tab:app_datasets}). We collect data with random policies only (no expert trajectories). The torques are: cartpole, Ornstein--Uhlenbeck noise ($\theta=0.2$, $\sigma=0.35$) clipped to [-1, 1], pushing the cart back beyond $\pm$1.6 m; pendulum, Ornstein--Uhlenbeck noise ($\theta=0.2$, $\sigma=0.35$) clipped to [-1, 1]; reacher, uniform random torques in [-1, 1]. Episodes start from: cartpole, uniform (position $\pm$1.8 m, any angle, velocities $\pm$5 m/s and $\pm$10 rad/s); pendulum, a uniform angle and an angular velocity uniform in $\pm$7 rad/s; reacher, the environment default, with the target not rendered.

\begin{table}[ht]
\centering
\small
\caption{Pixel datasets. Rate: frames per second of simulated time (one learned model step per frame).}\label{tab:app_datasets}
\begin{adjustbox}{max width=\linewidth}
\begin{tabular}{llrrrr}
\toprule
\textbf{Environment} & \textbf{Frames} & \textbf{Episodes $\times$ length} & \textbf{Rate} & \textbf{Train / val / test episodes} \\
\midrule
cartpole  & 200,000 & 2000 $\times$ 100 & 50 Hz & 1400 / 300 / 300 \\
pendulum  & 200,000 & 2000 $\times$ 100 & 25 Hz & 1400 / 300 / 300 \\
reacher   & 200,000 & 2000 $\times$ 100 & 8.3 Hz & 1400 / 300 / 300 \\
reacher   & 200,000 & 2000 $\times$ 100 & 25 Hz & 1400 / 300 / 300 \\
\bottomrule
\end{tabular}
\end{adjustbox}
\end{table}

\clearpage
\section{Baselines}\label{app:baselines}

We reproduced TD-MPC2 \citeref{tdmpc2}, DreamerV3 \citeref{dreamer} and LeWM \citep{lewm} from the authors' implementations and default configurations (four seeds each). TD-MPC2 and DreamerV3 use 64$\times$64 frames, learn online for up to one million environment steps with their original optimisation settings and no per-task tuning, and are evaluated with their learned controllers.

LeWM uses 224$\times$224 frames, rendered at this resolution, with a three-frame context. We keep the original LeWM configuration, including a frameskip of 5, so that on 25\,Hz data the model runs at 5\,Hz. On Pendulum and CartPole, LeWM is trained on the same datasets as \wv{} (\cref{app:dataset_collection}). On Reacher, we train two versions. LeWM (original data) uses the dataset released with LeWM, collected with random actions at an action repeat of 2 (25\,Hz) without the target rendered \citep{lewm}. LeWM (our data) uses our 25\,Hz Reacher dataset. Both share the same model step. LeWM is trained for 40 epochs with the SIGReg weight selected per environment (0.001 for CartPole, 0.1 for Pendulum and Reacher), and is controlled with its own CEM planner. Rewards are the sum of the native rewards over 500-step episodes. \Cref{tab:main,tab:reacher-timing} use LeWM (original data) for Reacher, and \cref{tab:lewm-protocol} compares both versions.

All baselines are evaluated on the test episodes of our datasets. For LeWM (original data), this is a shift away from its training distribution, which explains its low LRC and EQA on Reacher.

\paragraph{Rollout metrics and planning decouple for LeWM.}
Trained on the same Reacher data as \wv{}, LeWM (our data) reaches a 1\,s LRC of 95.2\% and an EQA of 85.6\%, yet obtains a reward of 21 and reaches only 14.7\% of goals 25 steps ahead (\cref{tab:lewm-protocol}). Since both LeWM versions share similar data-collection policy, model step and configuration, we attribute this gap to generalization: LRC and EQA are measured on held-out random-action episodes, whereas CEM evaluates action sequences and reaches states that are rare in this distribution, where the predictor can extrapolate poorly. Conversely, LeWM (original data) transfers poorly to our test split (LRC 0.0\%) but plans better under the original protocol (81.6\% success). For \wv{}, rollout metrics and planning agree, which we hypothesize is due to the vector-field parameterization and multi-step training.

\begin{table}[t]
    \centering
    \caption{\textbf{LeWM vs world2vec on Reacher.} Reacher only. Cons. / LRC / EQA (\%) on the test split as in Table 1: Cons. is the affine $R^2$ between seeds of one method on held-out episodes; LRC is the variance-weighted $R^2$ of the predicted against the encoded true latent; EQA is top-1 retrieval among 512 candidates. Reward is the Table 1 best-planner dm\_control return (maximum 1000). Success (\%) and planning time (s / episode): goal 25 or 200 dataset steps ahead, 2 s budget, mean over MPC seeds then bootstrap SE over training seeds. \WV{} 8.3 Hz is trained on the 8.3 Hz data and evaluated on the 25 Hz data with action averaging. Best per column in bold, second-best underlined (lower is better for s/ep).}
    \label{tab:lewm-protocol}
    \small
    \setlength{\tabcolsep}{3.5pt}
    \begin{tabular}{@{}lrrrrrrr@{}}
    \toprule
    & & & & & \multicolumn{2}{c}{Success} & \\
    \cmidrule(lr){6-7}
    Method & Cons. & LRC$_{1\,\mathrm{s}}$ & EQA$_{1\,\mathrm{s}}$ & Reward & 25 & 200 & s/ep \\
    \midrule
    LeWM (original data) & \underline{99.4} & 0.0 & \meanstd{4.7}{1.0} & \meanstd{69}{4} & \meanstd{81.6}{2.4} & \meanstd{52.9}{3.8} & \meanstd{8.64}{0.12} \\
    LeWM (our data) & \won{99.7} & \meanstd{95.2}{0.3} & \meanstd{85.6}{0.6} & \meanstd{21}{4} & \meanstd{14.7}{1.4} & \meanstd{2.4}{0.2} & \meanstd{8.32}{0.03} \\
    \midrule
    \wvsigreg{} 25\,Hz & 96.3 & \won{\meanstd{99.0}{0.1}} & \underline{\meanstd{96.3}{0.4}} & \underline{\meanstd{759}{10}} & \underline{\meanstd{85.8}{1.0}} & \meanstd{72.4}{0.2} & \meanstd{2.61}{0.06} \\
    \wvsigreg{} 8.3\,Hz & 97.6 & \meanstd{95.6}{0.2} & \meanstd{77.6}{1.8} & \meanstd{570}{20} & \meanstd{78.9}{1.4} & \meanstd{76.0}{2.2} & \underline{\meanstd{1.08}{0.07}} \\
    \wvdcl{} 25\,Hz & 98.5 & \won{\meanstd{99.0}{0.0}} & \won{\meanstd{97.3}{0.2}} & \meanstd{618}{4} & \won{\meanstd{86.2}{1.1}} & \won{\meanstd{80.2}{1.6}} & \meanstd{2.55}{0.03} \\
    \wvdcl{} 8.3\,Hz & 95.1 & \underline{\meanstd{95.7}{0.1}} & \meanstd{85.4}{1.4} & \won{\meanstd{895}{16}} & \meanstd{75.8}{2.7} & \underline{\meanstd{77.6}{2.8}} & \won{\meanstd{1.00}{0.01}} \\
    \bottomrule
    \end{tabular}
    \end{table}

\clearpage
\section{Experimental Details: Symbolic Regression with \wv{}} \label{app:sindy}

    \paragraph{Datasets.}
    The learned latent dynamics are evaluated on 71 systems from the \texttt{dysts} collection \cite{dysts}. For each system, we generate 1,000 ground-truth trajectories of 1,000 steps using a fourth-order Runge--Kutta (RK4) integrator with a step size $\Delta t_s$ equal to ten times the system-specific value defined by \citet{dysts}. To choose the initial conditions, we first simulate a noise-free, 1,000-step reference trajectory from the system's canonical initial state, then discard the first 100 states, determine the minimum and maximum values reached by each state coordinate over the remaining trajectory, and draw each coordinate of a new initial condition uniformly between these bounds. Trajectories containing nonfinite states or exceeding $10^6$ in absolute value are resampled.
    After each RK4 step, independent Gaussian noise is added to the simulated physical state: $\vs_{t+1}=\rvPhi_{\mathrm{RK4}}(\vs_t)+\bm{\epsilon}_t$. For state variable $i\in\{1,\ldots,d\}$, the noise has standard deviation $\alpha\sigma_{s,i}\sqrt{\Delta t_s/T_s}$, where $\sigma_{s,i}$ is that variable's attractor standard deviation reported in the \texttt{dysts} metadata and $T_s$ is the system's reference period. We use $\alpha\in\{0,0.1,0.5,1.0\}$, with $\alpha=0$ giving noise-free trajectories.
    This is followed by applying a fixed, \textit{injective} mixing function $\rvg:\mathbb{R}^{d}\to\mathbb{R}^{n}$ to the true latent trajectory $\vs\in\mathbb{R}^d$, yielding observed time series $\vx=\rvg(\vs)\in\mathbb{R}^{n}$. Following \citet{zimmermann_contrastive_2021}, the mixing function consists of a fixed four-layer MLP followed by a linear projection to $n = 50$ dimensions. Entire trajectories are assigned to training, validation, and test sets in a 70:15:15 ratio. All models use a three-layer MLP encoder with 128 hidden units, trained with the one-step \wv{} objective.

    \paragraph{Models.}
    Three dynamics models are evaluated on the generated datasets. Identity dynamics provide a baseline without a learned dynamics model: $\hat{\vz}_{t+1}=\vz_t$. The MLP dynamics model parameterizes a continuous-time flow field, $\dot{\vz}=\hat{\rvf}^{\mathrm{MLP}}(\vz)$, whose one-step prediction is obtained by RK4 integration. Finally, the SINDy model represents the flow field as $\dot{\vz}=\boldsymbol{\Xi}\,\boldsymbol{\theta}(\vz)$, where $\boldsymbol{\theta}(\vz)\in\mathbb{R}^{p}$ contains all monomials of $\vz$ up to degree $q$, including the constant, with the \texttt{dysts} system-specific maximum degree $q\in\{2,3\}$ being assumed to be known, and $\boldsymbol{\Xi}\in\mathbb{R}^{d\times p}$ is a sparse coefficient matrix whose $k$-th row gives the governing equation of $z_k$. Finally, an $L_1$ penalty, $\lambda_{\mathrm{SINDy}}\|\boldsymbol{\Xi}\|_1$ with $\lambda_{\mathrm{SINDy}}=10^{-5}$, encourages sparse coefficients, and (unlike for other SINDy implementations \cite{sindy, sindy_ae, sindy_bae}), no coefficient thresholding is applied during training.
    All models are trained for 10,000 iterations, and we tune only the weight of the DCL Regularizer (\cref{app:objectives}) over a grid of $\lambda \in\{0.01,0.1,0.5,1.0\}$ (\cref{eq:rollout}). During analysis, for all systems, (except for the qualitative rollout plots), we select the ensemble across uniformity weights that has the highest mean $\mathrm{LRC}_{50}$ on the validation set and then report the results on the validation set.

    \paragraph{Evaluation.}
    The performance of the resulting models is assessed across four metrics. Affine recovery $R^2$ measures how well the physical state can be recovered from the learned representation. Consistency quantifies mean pairwise affine agreement between representations learned from three different model seeds. iDyn$R^2_{50}$ compares ground-truth and learned 50-step rollouts after affine alignment, while $\mathrm{LRC}_{50}$ compares learned rollouts with encodings of the observed future.
    
    \paragraph{Symbolic regression.}
    For qualitative inspection of rollouts and recovered equations in \cref{fig:sindy}, we first rank the recovered SINDy systems by $\mathrm{LRC}_{50}$ and retain only the top ten systems. Among these, we select the system whose learned SINDy models are most consistent across seeds. Because each seed learns its own latent coordinate system, we first express all three SINDy models in a common space, choosing the seed \(r\) with the highest validation \(\mathrm{LRC}_{50}\) as the reference. For each seed \(j \neq r\), we fit the affine transformation \(\vz^{(r)} \approx \mL_{j\to r}\vz^{(j)} + \vb_{j\to r}\) by least squares on paired embeddings.
    Since \(\dot{\vz}^{(r)}=\mL_{j\to r}\dot{\vz}^{(j)}\), the SINDy coefficient matrix learned by seed \(j\) becomes \(\boldsymbol{\Xi}_{j\to r} = \mL_{j\to r}\boldsymbol{\Xi}_j\mM_{j\to r}\) in the reference coordinates, where \(\mM_{j\to r}\in\mathbb{R}^{p\times p}\) describes how the polynomial features transform under the affine coordinate change\citep{dysco}, \(\boldsymbol{\theta}\big(\mL_{j\to r}^{-1}(\vz^{(r)}-\vb_{j\to r})\big) = \mM_{j\to r}\boldsymbol{\theta}(\vz^{(r)})\). The reference model remains unchanged, \(\boldsymbol{\Xi}_{r\to r}=\boldsymbol{\Xi}_r\).
    Once all three models are expressed in the same coordinate system, we quantify their agreement using the mean pairwise Frobenius distance between their coefficient matrices, \(D = \frac{1}{3}\sum_{1\leq j<k\leq 3}\left\|\boldsymbol{\Xi}_{j\to r}-\boldsymbol{\Xi}_{k\to r}\right\|_F\). Among the candidate systems, we select the one with the smallest \(D\), corresponding to the SINDy models with the highest agreement across seeds. Importantly, both the alignment and the coefficient-distance criterion depend only on the learned embeddings and the learned SINDy coefficients and therefore do not require access to the ground-truth latent states or dynamics.

\end{document}